%% file: main.tex
\documentclass[pagebackref,breaklinks,colorlinks,allcolors=iccvblue]{fairmeta}

\definecolor{iccvblue}{rgb}{0.21,0.49,0.74}

\usepackage{url}
\usepackage{makecell}

\usepackage{epsfig}
\usepackage{graphicx}
\usepackage{amsmath}
\usepackage{amssymb}
\usepackage{fontawesome5}
\usepackage{graphicx}
\usepackage{amsmath}
\usepackage{amssymb}
\usepackage{booktabs}
\usepackage[ruled,vlined]{algorithm2e}
\usepackage{algorithmic}
\usepackage{colortbl} % for coloring cells
\usepackage{booktabs} % for horizontal rules
\usepackage{multirow, booktabs, graphicx, xcolor, colortbl, arydshln, tikz}
\usetikzlibrary{tikzmark}
\usepackage{makecell}
\usepackage{siunitx} % for 'S' column type
\usepackage{mdframed}
\usepackage{tablefootnote} 

\usepackage{wrapfig}

\newcolumntype{g}{>{\columncolor{gray!30}}c}

\usepackage{tcolorbox}
\tcbuselibrary{most}

\definecolor{citecolor}{HTML}{0071bc}
\hypersetup{
    colorlinks=true,%
    citecolor=citecolor,%
    filecolor=citecolor,%
    linkcolor=citecolor,%
    urlcolor=citecolor
}

\tcbuselibrary{skins}

\definecolor{userbg}{RGB}{245, 245, 245}
\definecolor{userborder}{RGB}{210, 229, 255}
\definecolor{userfont}{RGB}{0, 0, 0}

\tcbset{
  enhanced,
  boxrule=1pt,
  colback=userbg,
  colframe=userborder,
  fonttitle=\bfseries,
  coltitle=userfont,
  boxsep=5pt,
  arc=5pt,
  outer arc=5pt,
  left=10pt,
  right=10pt,
  top=2pt,
  bottom=2pt,
  attach boxed title to top left={yshift=-0.25\baselineskip-1mm,xshift=2mm},
  boxed title style={
    colback=userborder,
    sharp corners,
    rounded corners=northeast,
    arc=3pt,
    outer arc=3pt,
    boxrule=0pt,
    boxsep=2pt,
  },
}
\usepackage{pgf}
\usepackage{adjustbox}

\usepackage{multirow}
\usepackage{colortbl} % For color
\usepackage{booktabs}
\usepackage{tcolorbox} % for colored description boxes
\usepackage{enumitem}
\usepackage{graphicx}
\usepackage{multirow}
\usepackage{array}
\definecolor{listcolor}{RGB}{50,120,230}
\usepackage{arydshln}
\usepackage{url}
\usepackage{caption}

\newcounter{researchquestion}

\newcommand{\researchquestion}[2][]{%
  \vspace{0.8em}% Add space before
  \refstepcounter{researchquestion}% Increment counter
  \begin{tcolorbox}[
    enhanced,
    colback=blue!5,
    colframe=blue!70!black,
    fonttitle={\fontsize{10.5pt}{12.8pt}\selectfont\bfseries\color{blue!20!black}},  % Custom size between \normalsize and \large
    title=Question \theresearchquestion,
    toprule=1.5pt,
    bottomrule=0.8pt,
    leftrule=0.8pt,
    rightrule=0.8pt,
    left=6pt,
    right=6pt,
    top=6pt,
    bottom=6pt,
    boxsep=3pt
  ]
  \normalsize #2
  \end{tcolorbox}
  \ifx\\#1\\\else\label{rq:#1}\fi% Add label if provided
  \vspace{0.5em}% Add space after
}

\title{HazyDet: Open-Source Benchmark for Drone-View Object Detection with Depth-Cues in Hazy Scenes}

\author[1]{Changfeng Feng}
\author[2]{Zhenyuan Chen}
\author[2,3]{Xiang Li}
\author[1]{Chunping Wang}
\author[2]{Jian Yang}
\author[2,3]{Ming-Ming Cheng}
\author[2,3,\dagger]{Yimian Dai}
\author[1,\dagger]{Qiang Fu}

\affiliation[1]{Army Engineering University}
\affiliation[2]{VCIP, School of Computer Science, Nankai University}
\affiliation[3]{NKIARI, Shenzhen Futian}
\contribution[\dagger]{Corresponding Author}

\abstract{
Object detection from aerial platforms under adverse atmospheric conditions, particularly haze, is paramount for robust drone autonomy.
Yet, this domain remains largely underexplored, primarily hindered by the absence of specialized benchmarks.
To bridge this gap, we present \textit{HazyDet}, the first, large-scale benchmark specifically designed for drone-view object detection in hazy conditions.
Comprising 383,000 real-world instances derived from both naturally hazy captures and synthetically hazed scenes augmented from clear images, HazyDet provides a challenging and realistic testbed for advancing detection algorithms. 
To address the severe visual degradation induced by haze, we propose the Depth-Conditioned Detector (DeCoDet), a novel architecture that integrates a Depth-Conditioned Kernel to dynamically modulate feature representations based on depth cues.
The practical efficacy and robustness of DeCoDet are further enhanced by its training with a Progressive Domain Fine-Tuning (PDFT) strategy to navigate synthetic-to-real domain shifts, and a Scale-Invariant Refurbishment Loss (SIRLoss) to ensure resilient learning from potentially noisy depth annotations.
Comprehensive empirical validation on HazyDet substantiates the superiority of our unified DeCoDet framework, which achieves state-of-the-art performance, surpassing the closest competitor by a notable +1.5\% mAP on challenging real-world hazy test scenarios.
Our dataset and toolkit are available at \url{https://github.com/GrokCV/HazyDet}.
}

\date{May 25, 2025}

\project{\href{https://github.com/GrokCV/HazyDet}{\sffamily \fontsize{8.8pt}{11pt}\selectfont \texttt{https://github.com/GrokCV/HazyDet}}}

\begin{document}
\maketitle

\input{arXiv-contents/introduction}

\input{arXiv-contents/related-work}
\input{arXiv-contents/dataset}

\input{arXiv-contents/method}

\input{arXiv-contents/experiment}
\input{arXiv-contents/conclusion}

\definecolor{blue3}{HTML}{5D8DFD}
\definecolor{green3}{HTML}{88B06D}
\definecolor{orange3}{HTML}{F5A83D}
\definecolor{red3}{HTML}{F5433D}
\definecolor{blue1}{HTML}{AEC6FE}
\definecolor{green1}{HTML}{B3CDA2}
\definecolor{orange1}{HTML}{F9CB8A}
\definecolor{red1}{HTML}{FBA09D}
\definecolor{lightgray}{gray}{1.0}
\definecolor{cambriangray}{gray}{0.9}

\section{Acknowledgements}
\label{sec:acknowledgements}
% The authors would like to thank the editor and the anonymous reviewers for their critical and constructive comments and suggestions.
The authors would like to thank the editor and the anonymous reviewers for their critical and constructive comments and suggestions.
This research is supported by the National Natural Science Foundation of China (Grant No. 62301261, 62206134, U24A20330, 62361166670) and the Fellowship of China Postdoctoral Science Foundation (Grant No. 2021M701727).
We acknowledge the Tianjin Key Laboratory of Visual Computing and Intelligent Perception (VCIP) for their essential resources. 
We extend our profound gratitude to Professor Pengfei Zhu and the dedicated AISKYEYE team from Tianjin University, whose invaluable data support has been instrumental to our research.

\clearpage

%%% References
{
    \small
    \bibliographystyle{ieeenat_fullname}
    \bibliography{main}
}

\input{appendix.tex}

\end{document}

%% file: arXiv-contents/introduction.tex
\section{Introduction} 
\label{sec:introduction}

\begin{wrapfigure}{r}{0.453\textwidth}
    \vspace{-1.5em}
    \centering
    % \captionsetup{font={scriptsize}}
    \includegraphics[width=0.47\textwidth]{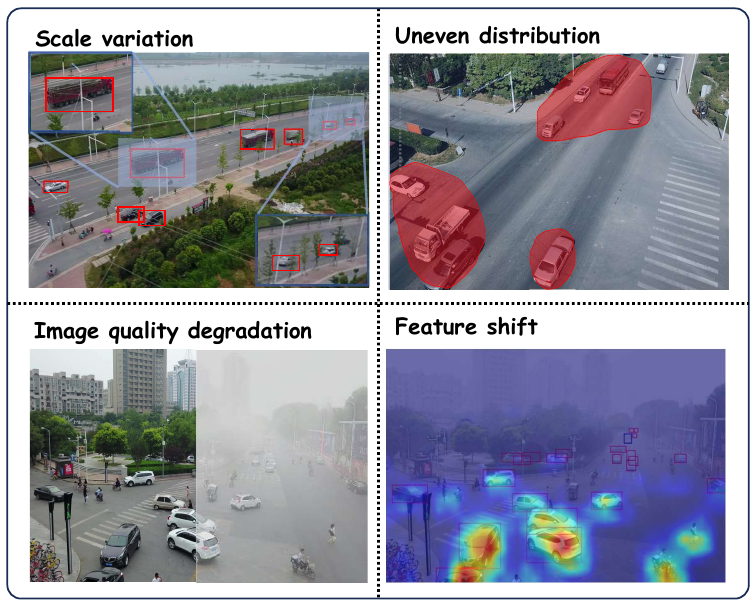}
    % \vspace{-2pt}
    \caption{\textbf{Challenges faced by drone-view object detection under hazy conditions.}}
    \label{fig:challenges}  
    \vspace{-2em}
\end{wrapfigure} 

Unmanned Aerial Vehicles (UAVs), or drones, have rapidly become indispensable tools in a diverse range of applications, due to their low cost and flexibility.

This widespread adoption is fundamentally underpinned by the onboard visual perception, with object detection serving as a vital component for environmental understanding.
Despite promising progress in recent years, ensuring reliable perception in challenging real-world conditions remains a significant hurdle.
Among various factors, adverse atmospheric effects, particularly haze, present a persistent obstacle that greatly undermines the robustness of drone-based detection systems.

\begin{figure*}[th]
   % \vspace{-2em}
   \centering   
   \includegraphics[width=1\linewidth]{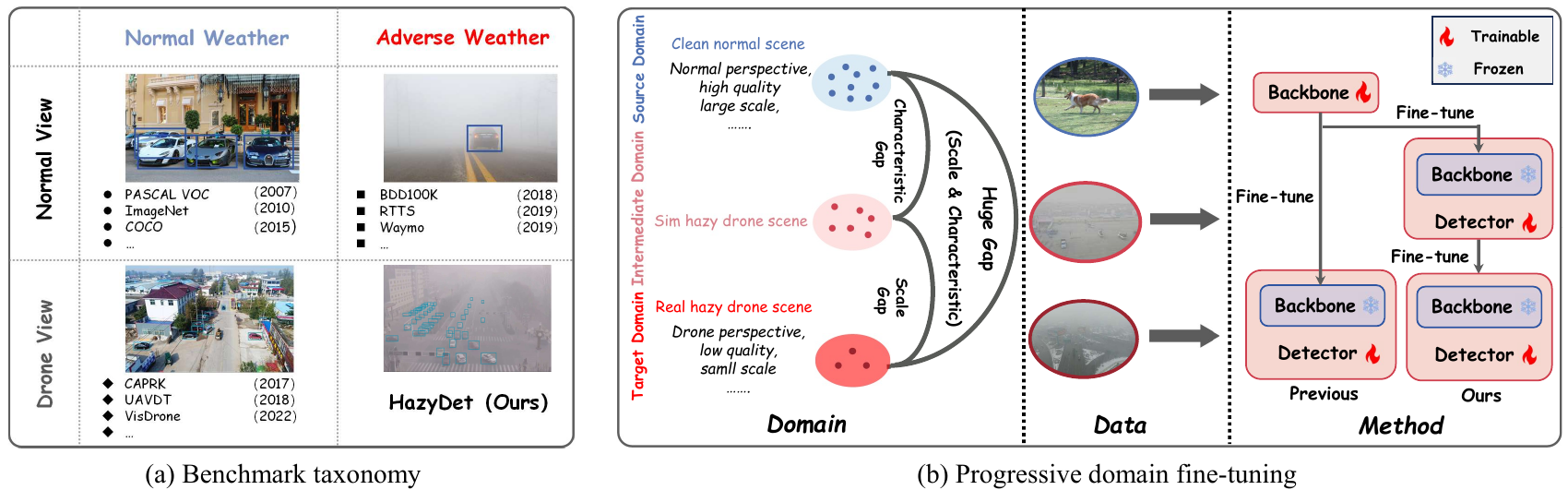}
   % \captionsetup{font={footnotesize}}
   \caption{\textbf{Data Landscape and Domain Fine-Tuning Strategy.} (a) HazyDet uniquely addresses the critical gap in drone-view adverse-weather scenarios.
   (b) PDFT paradigm bridges the simulated-to-real domain gap.}  
   \label{fig:dataset and PDFT}  
    % \vspace{-2em}
    \vspace{-1.5\baselineskip}
\end{figure*}

Even in clear weather, as shown in Fig.~\ref{fig:challenges} (top row), object detection from drones presents unique difficulties compared to ground-level vision \citep{CVPR2022QueryDet,CVPR2023CEASC}.
Most prior efforts in UAV object detection have focused on the intrinsic complexities of aerial imagery:
\textit{(1) Scale Variation}: aerial images contain objects at varying scales due to changing altitudes and viewpoints, with small objects predominating. Researchers address this through multi-scale feature integration using hierarchical networks to aggregate information across different spatial resolutions \citep{CVPR2023CEASC,TIM2024MFFSODNet}.
\textit{(2) Non-uniform Spatial Distribution}: objects tend to be scattered irregularly across the scene, unlike the more centered compositions often found in standard datasets.
To this end, others have explored coarse-to-fine detection pipelines, where initial region proposals are progressively refined to achieve precise localization \citep{TIP2021GLSAN,TMM2023AdaZoom}.
These advances may have robust results under normal conditions, yet their effectiveness in the presence of environmental degradations is far less understood.

In practice, UAVs are mostly deployed in outdoor environments where weather and atmosphere can vary dramatically.
Haze is especially common, and it introduces two additional challenges: (1) reduced image quality due to diminished contrast and color shifts, (2) and a more subtle, yet equally problematic, shift in the underlying feature distribution that models rely on for recognition. While some research, mainly from the autonomous driving field, has tried to address weather effects by combining detection with image restoration \citep{AAAI2020FFANet,TPAMI2020DSNet,AAAI2022IAYOLO,TPAMI2023DetectionFriendlyDehazing}, these solutions are not transferable to UAVs due to their reliance on ground-level priors and the conspicuous absence of drone-specific hazy benchmarks (Fig.~\ref{fig:dataset and PDFT} (a)). Furthermore, image restoration can sometimes create artifacts that actually harm detection performance \citep{TPAMI2022AttackNetwork}. Critically, a fundamental oversight we identified is that most current methods don't effectively use depth information.
This is a pivotal cue, given that \textit{object scale is intrinsically tied to depth in aerial views, and haze density is also physically governed by it}.

In this paper, we propose \textbf{\textit{HazyDet}}, the first large-scale benchmark dataset created specifically for drone-view object detection in hazy conditions, containing 383,000 instances from real-world hazy images and high-fidelity synthetic scenes. Motivated by the depth-dependent nature of haze degradation,
we further propose the \textbf{De}pth-\textbf{Co}nditioned \textbf{Det}ector (\textbf{DeCoDet}), a novel framework that eschews explicit dehazing and instead leverages a dynamic Depth-Conditioned Kernel to adapt feature representations based on estimated depth cues. To address the distributional shift between clear, synthetic haze, and real-world haze domains, DeCoDet is trained using a Progressive Domain Fine-Tuning (PDFT) strategy jointly with a Scale-Invariant Refurbishment Loss (SIRLoss), enabling robust depth-aware learning even from noisy pseudo annotations. This unified framework achieves state-of-the-art performance on challenging real-world hazy drone-view scenarios, surpassing the second-best method by a notable margin of +1.5\% mAP.

In summary, our contributions are threefold:

\begin{itemize}[itemsep=3pt, topsep=0pt, parsep=0pt, partopsep=0pt]

   \item \textbf{HazyDet Benchmark}: We introduce the first large-scale benchmark dataset for drone-view object detection in hazy conditions, containing 383,000 instances.
    
    \item \textbf{DeCoDet Framework}: We propose DeCoDet, a novel detection framework that dynamically leverages depth cues in hazy aerial imagery as well as PDFT strategy and SIRLoss, ensuring robust depth-aware learning across domain shifts and from noisy depth supervision.
    
    \item \textbf{State-of-the-Art Performance}: Our DeCoDet achieves a 
    notable +1.5\% mAP improvement over the closest competitor on  real-world hazy test-set.

\end{itemize}

%% file: arXiv-contents/related-work.tex
\section{Related Work} \label{sec:related}

\vspace{-1\baselineskip}
\input{tables/datasets_compare}

\noindent\textbf{Drone-View Detection \& Adverse Weather Restoration Datasets.}
Over the past decade, as shown in Tab.~\ref{tab:dataset_comparison},
several benchmarks, such as VEDAI~\citep{JVCIR2016VEDAI}, CARPK~\citep{ICCV2017CAPRK}, UAVDT~\citep{ECCV2018UAVDT}, and VisDrone~\citep{TPAMI2022VisDrone}, have become foundational to drone-view object detection research, offering diverse settings and rich annotations. However, a salient limitation pervades these existing resources: they are predominantly constructed under clear atmospheric conditions, offering little insight into real-world performance when drones operate in degraded environments. In parallel, the image restoration community has curated a range of dehazing and foggy datasets~\citep{ICIP2016Dhzae,ICIP2019DenseHazes,TIP2019RESIDE,CVPR2021_4KDehaze,TIP2023RSHaze,TMM2023RW-HAZE}, some even capturing authentic haze.
However, these datasets are invariably designed for image restoration tasks and lack the comprehensive detection annotations or the aerial perspectives required for advancing high-level vision in the drone domain.
Notably, while datasets such as RTTS~\citep{TIP2019RESIDE} and A2I2~\citep{TIP2023A2I2} have introduced semantic annotations for foggy imagery, they are either confined to non-aerial perspectives or constrained by limited scale and reliance on artificial smoke.
To the best of our knowledge, HazyDet is the first dataset to bridge this divide, providing a large-scale, meticulously labeled benchmark for drone-view object detection in hazy scenes.

\vspace{-0.3\baselineskip}

\noindent\textbf{Drone-View Object Detection Methods.}
To address the variation in object scales, existing approaches have converged on multi-scale feature fusion.
For instance, CFANet \citep{TGRS2023CFANet} employs cross-layer feature aggregation to sharpen the detection of small, scale-diverse targets, while MFFSODNet \citep{TIM2024MFFSODNet} implements multi-branch convolutions with varied kernel sizes to capture features across multiple receptive fields.
Beyond scale, the non-uniform spatial distribution of objects in drone imagery further necessitates a series of coarse-to-fine pipelines.
To this end, GLSANet \citep{TIP2021GLSAN} incorporates a self-adaptive region selection mechanism coupled with local super-resolution to refine densely populated areas, while UFPMP-DET \citep{AAAI2022UFPMPDet} integrates a unified foreground packing strategy with multi-proxy learning to address sparsely distributed targets. ClusDet \citep{ICCV2019ClusterDet} attempts to predict and adaptively resize clustered object regions.
Despite these advances, the question of how adverse weather conditions fundamentally influence the detection capabilities of UAVs remains largely unresolved.

\vspace{-0.3\baselineskip}

\noindent\textbf{Object Detection in Adverse Conditions.}
Object detection in adverse environments presents substantial complexities beyond clear-weather scenarios, primarily due to degraded image fidelity and atypical visual cues \citep{TGRS2024Night-timedrone,TIV2024MisalignedMMDrone}. 
Methods addressing this typically adopt either separate or joint optimization.
Separate paradigms first employ restoration algorithms to enhance image quality before detection; however, this often yields insufficient improvement in detection accuracy and can detrimentally suppress high-frequency details crucial for discerning small aerial targets \citep{TIP2020ConnectingDenoisingwitHighVision}.
Conversely, joint optimization frameworks integrate restoration and detection, such as AOD-Net \citep{ICCV2017AODNet}, IA-YOLO \citep{AAAI2022IAYOLO}, DSNet \citep{TPAMI2020DSNet}, and BAD-Net \citep{TPAMI2023DetectionFriendlyDehazing}. Yet, these approaches frequently contend with conflicting optimization objectives and a strong dependency on meticulously paired data, which is often scarce.

\vspace{-0.3\baselineskip}

Our DeCoDet distinguishes itself by leveraging auxiliary depth information rather than directly coupling detection with image restoration. This strategy enhances perceptual robustness in challenging weather conditions while obviating the need for paired datasets, leading to significant improvements in detection performance within hazy environments.

%% file: tables/datasets_compare.tex
\begin{table}[ht]
\centering
\caption{Comparative landscape of relevant datasets. This positions HazyDet as the first large-scale (383K instances) drone-view benchmark for object detection under hazy conditions, uniquely providing extensive real-world and high-fidelity simulated scenes.}
\vspace{-0.5em}
\label{tab:dataset_comparison}
\setlength{\tabcolsep}{8pt}
\begin{tabular}{>{\raggedright}p{2.8cm}cccccccc} % 移除@{}避免背景色溢出
  \toprule
  \textbf{Dataset} & \textbf{Drone-View} & \textbf{Detection} & \textbf{Hazy} & \textbf{Real} & \textbf{Sim.} & \textbf{\#Inst.} & \textbf{\#Img} & \textbf{Venue} \\
  \midrule
D-HAZE & $\times$ & $\times$ & \checkmark & $\times$ & \checkmark & $\times$ & 1,449 & ICIP'16 \\
Dense-HAZE & $\times$ & $\times$ & \checkmark & $\times$ & \checkmark & $\times$ & 95 & ICIP'19 \\
RESIDE & $\times$ & $\times$ & \checkmark & $\times$ & \checkmark & $\times$ & 13,990 & TIP'19 \\
4KDehaze & $\times$ & $\times$ & \checkmark & $\times$ & \checkmark & $\times$ & 8,200 & CVPR'21 \\
RS-Haze & $\times$ & $\times$ & \checkmark & $\times$ & \checkmark & $\times$ & 50k & TIP'23 \\
RW-Haze & $\times$ & $\times$ & \checkmark & \checkmark & $\times$ & $\times$ & 210 & TMM'23 \\
RTTS & $\times$ & \checkmark & \checkmark & \checkmark & $\times$ & 41k & 4,322 & TIP'19 \\
CAPRK & \checkmark & \checkmark & $\times$ & $\times$ & $\times$ & 89k & 1,448 & ICCV'17 \\
UAVDT & \checkmark & \checkmark & $\times$ & $\times$ & $\times$ & 842k & 37k & ECCV'18 \\
VisDrone & \checkmark & \checkmark & $\times$ & $\times$ & $\times$ & 343k & 10k & TPAMI'22 \\
A2I2 & \checkmark & \checkmark & \checkmark & \checkmark & $\times$ & 3,898 & 359 & TIP'23 \\
  \rowcolor{gray!15}
  \textbf{HazyDet (Ours)} & \checkmark & \checkmark & \checkmark & \checkmark & \checkmark & \textbf{383k} & \textbf{11.6k} & \textbf{---} \\
  \bottomrule
\end{tabular}
\end{table}

%% file: arXiv-contents/dataset.tex
\section{HazyDet: A New Benchmark Dataset} \label{sec:dataset}

\subsection{Dataset Construction}

\begin{figure}[ht]
    \centering
    \includegraphics[width=0.95\linewidth]{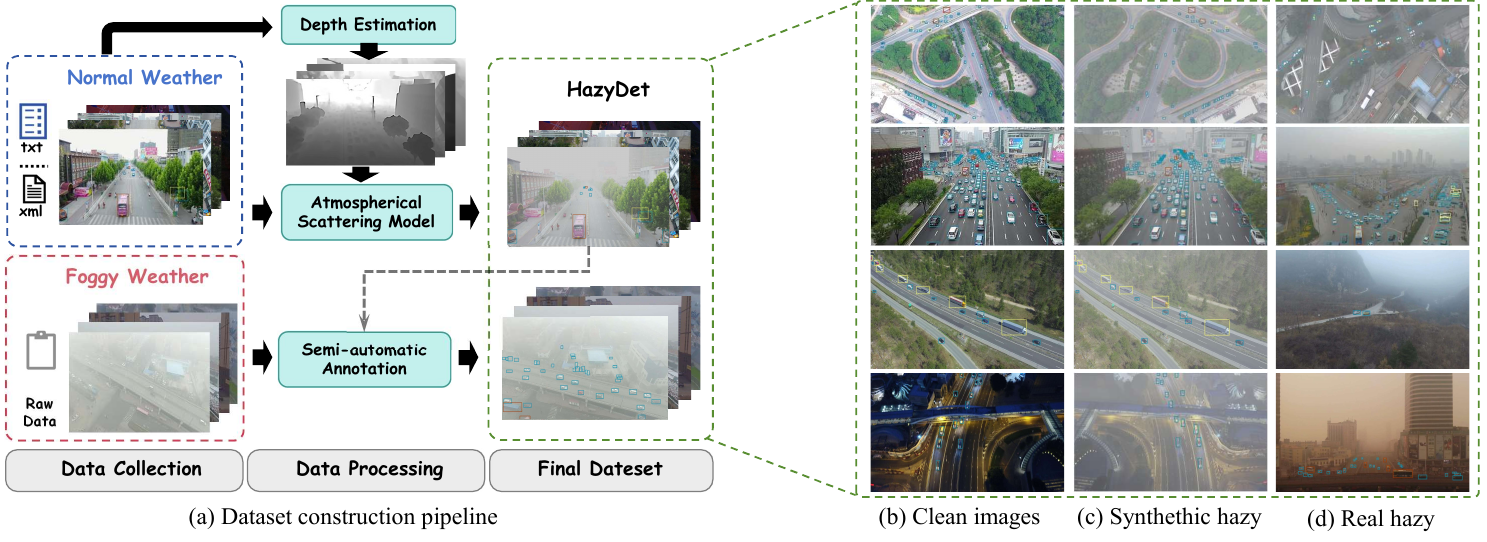}
    % \captionsetup{font={scriptsize}}
    \vspace{-1em}
    \caption{   
    \textbf{Construction pipeline and representative samples of the HazyDet dataset.} (a) Simulated data is generated using physics-based simulation based on the Atmospheric Scattering Model (ASM), while semi-automatic annotation techniques are employed to improve the accuracy and efficiency of annotations for real-world data. (b) Clean images; (c) Synthetically generated hazy images; (d) Real-world hazy images captured in HazyDet.
    }
    \label{fig:dataset_pipeline_samples}
    % \vspace{-1em}
\end{figure}

\input{tables/dataset_count}

We adopt a dual-track strategy that combines physics-driven synthesis with field acquisition of real-world data (Fig.~\ref{fig:dataset_pipeline_samples}), addressing the scarcity of authentic hazy observations while leveraging simulated scene diversity. Stratified sampling is used to partition the datasets, with simulated data divided into training, validation, and test subsets at an 8:1:2 ratio, and real-world data split into training and test subsets at a 2:1 ratio. Detailed distribution statistics are provided in Tab.~\ref{tab:dataset_count}.

\textbf{Real hazy image acquisition.} To benchmark real-world performance, we capture 600 drone images under genuine fog across urban, rural, and littoral regions with diverse flight altitudes and viewing angles.

\textbf{Synthetic data generation.}
To synthesize hazy data, we employed the Atmospheric Scattering Model (ASM), a physics-driven framework defined as:
\begin{equation}
\label{eq1}
I(x,y)=J(x,y)t(x,y)+A(1-t(x,y)),t\left(x,y\right)=e^{-\beta d(x,y)}.
\end{equation}
where $I(x,y)$ denotes the observed hazy image, $J(x,y)$ the clear scene radiance, $A$ the global atmospheric light, $\beta$ the scattering coefficient, and $d(x,y)$ the scene depth. 
Departing from prior works that relied on fixed or broadly sampled parameters \citep{TIP2019RESIDE,TGRS2023Trinity-Net}, we model $A$ and $ \beta$ with truncated normal distributions: $A \sim \mathcal{N}(0.8, 0.05^2)$ in $[0.7, 0.9]$ and $ \beta \sim \mathcal{N}(0.045, 0.02^2)$ in $[0.02, 0.16]$, enhancing the realism of haze effects. Depth $d(x,y)$  is derived using advanced estimation models, ensuring fidelity in degradation simulation. To ensure the authenticity of the simulation in the depth estimation model selection, we organize a survey with 280 participants (as shown in Fig. \ref{fig:vote}).

\subsection{Dataset Statistics and Characteristics}

\textbf{Authenticity assessment.} The objective FID/KID scores \citep{arxiv2021FID,ICLR2018KID}, together with the subjective visual comparisons, consistently indicate that our synthetic images are perceptually and statistically closer to real drone haze than prior datasets such as RESIDE‐out \citep{TIP2019RESIDE} or 4KDehaze \citep{CVPR2021_4KDehaze} (Fig.~\ref{fig:authenticity}).

\textbf{Category balance and object scale.} The benchmark covers cars, trucks, buses, and miscellaneous traffic participants. As depicted in Fig.~\ref{fig:data_scale}, instance sizes span more than four octaves, with a pronounced long tail of small objects (64–256 px$^2$), mirroring practical drone missions.

\textbf{Depth–size correlation.} According to the ASM, distance amplifies the effect of fog on atmospheric transmission, making images more blurry, while the aerial viewpoint amplifies perspective cues: objects nearer the camera appear markedly larger. Quantitative analysis (Fig.~\ref{fig:depth_ana}) confirms a strong inverse correlation between estimated depth and bounding‐box area, reinforcing our architectural decision to condition detection on depth.

Relative to existing synthetic datasets, HazyDet exhibits lower FID/KID scores with respect to real haze, higher object density, and explicit depth annotations, constituting an indispensable testbed for designing haze-robust aerial detectors. As the first meticulously engineered, depth-aware benchmark for drone-view detection in haze, HazyDet supplies the community with a foundation for principled algorithmic progress and reproducible evaluation under genuinely challenging conditions.

%% file: tables/dataset_count.tex
\begin{wraptable}{rt}{0.5\linewidth}  
    \centering  
    % \captionsetup{font={scriptsize}}  
    \vspace{-2em}  
    \captionof{table}{  
        Statistics of images and instances across different dataset subsets.  
        Targets are divided into three size groups: small targets have an area-to-image-area  
        ratio of <0.1\%, medium targets cover 0.1\%--1\%, and large targets exceed 1\%.  
    }  
    \label{tab:dataset_count}  
    \renewcommand{\arraystretch}{0.95} % 缩小行间距
    \resizebox{0.99\linewidth}{!}{  
        \begin{tabular}{lccclccc}  
            \toprule  
            \multicolumn{1}{l}{Split} &  
            \multicolumn{1}{c}{\#Image} &  
            \multicolumn{1}{c}{\#Instance} &  
            \multicolumn{1}{c}{Class} &  
            \multicolumn{3}{c}{Object Size} \\
            \cmidrule(lr){5-7}  
            & & & & Small & Medium & Large \\
            \midrule  
            \multirow{3}{*}{Train} & \multirow{3}{*}{8,000} & \multirow{3}{*}{264,511}  
              & Car   & 159,491 & 77,527 & 5,177 \\
            & & & Truck & 4,197   & 6,262  & 1,167 \\
            & & & Bus   & 1,990   & 7,879  & 861   \\
            \midrule  
            \multirow{3}{*}{Val}   & \multirow{3}{*}{1,000} & \multirow{3}{*}{34,560}  
              & Car   & 21,051  & 9,881  & 630   \\
            & & & Truck & 552     & 853    & 103   \\
            & & & Bus   & 243     & 1,122  & 125   \\
            \midrule  
            \multirow{3}{*}{Test}  & \multirow{3}{*}{2,000} & \multirow{3}{*}{65,322}  
              & Car   & 38,910  & 19,860 & 1,256 \\
            & & & Truck & 881     & 1,409  & 263   \\
            & & & Bus   & 473     & 1,991  & 279   \\
            \midrule  
            \multirow{3}{*}{Real-world Train} & \multirow{3}{*}{400} & \multirow{3}{*}{13,753}  
              & Car   & 5,816   & 6,487  & 695 \\
            & & & Truck & 86     & 204    & 57    \\
            & & & Bus   & 52      & 256    & 100   \\
            \midrule  
            \multirow{3}{*}{Real-world Test} & \multirow{3}{*}{200} & \multirow{3}{*}{5,543}  
              & Car   & 2,351   & 2,506  & 365 \\
            & & & Truck & 26     & 86    & 30    \\
            & & & Bus   & 17      & 107    & 55   \\
            \bottomrule  
        \end{tabular}  
    }  
    \vspace{-1em}  
\end{wraptable}

%% file: arXiv-contents/method.tex
\vspace{-0.5\baselineskip}
\begin{figure*}[t]
    \centering
    \includegraphics[width=0.95\linewidth]{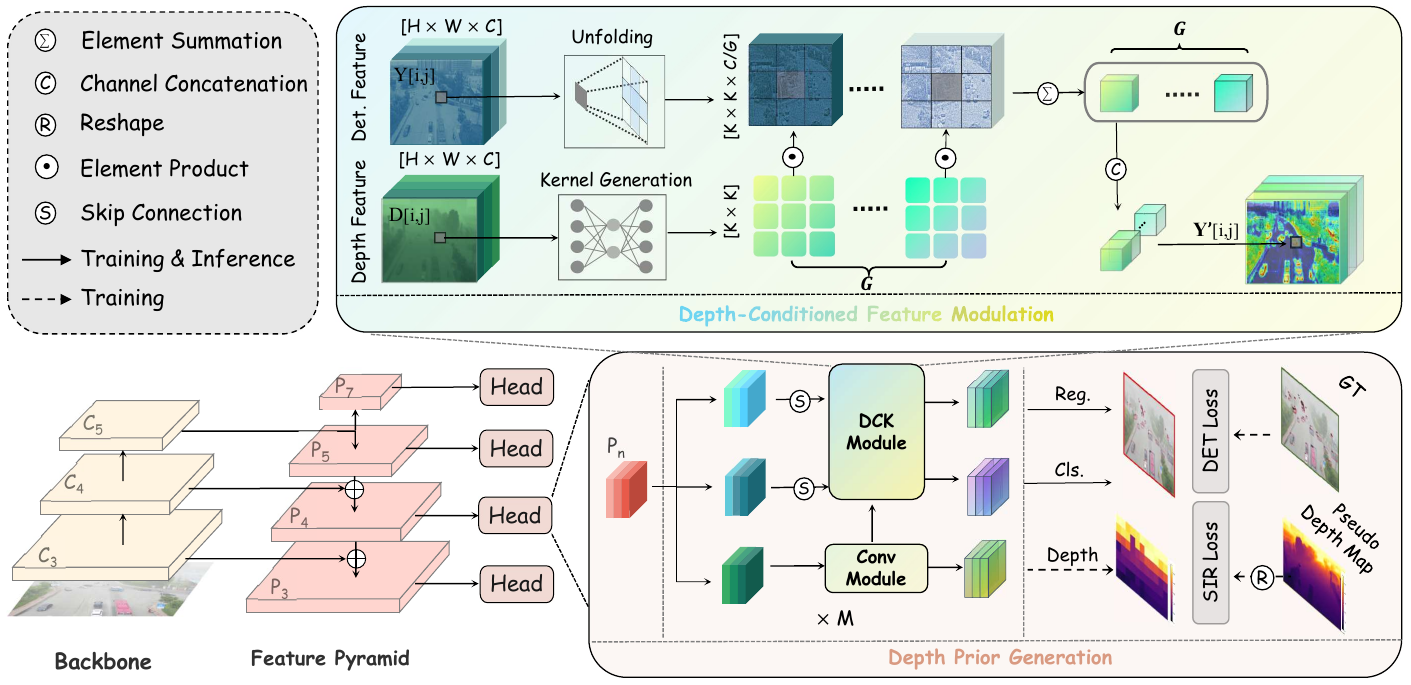}
    \vspace{-0.5\baselineskip}    
    % \captionsetup{font={scriptsize}}
    \caption{\textbf{The framework of DeCoDet.} The network comprises a backbone and a feature pyramid network for feature extraction. Each detection head incorporates depth-specific convolutional modules to infer depth maps at multiple scales, with the scale-invariant restoration loss (SIRLoss) computed with respect to pseudo-depth maps to facilitate the generation of depth priors. The intermediate depth features incorporating these depth priors are subsequently utilized by the Depth Conditioned Kernel (DCK) module to dynamically generate filter kernels, thereby adaptively modulating the detection features.}
    \label{fig:network}
    % \vspace{-2em}
    \vspace{-1\baselineskip}
\end{figure*}

\section{DeCoDet: Depth-Conditioned Detector} 
\label{sec:method}  
\vspace{-0.5\baselineskip}
\subsection{Depth-Aware Representation Learning}

% \label{subsec:MDDH}

At the core of our DeCoDet lies the insight that haze-induced visual ambiguities in aerial imagery are fundamentally governed by scene depth, which modulates both the severity of degradation and the apparent scale of objects. Rather than treating depth as an auxiliary prediction or an isolated task, we architect DeCoDet to treat depth as an intrinsic modality, seamlessly woven into the detection pipeline.
First, we develop a lightweight depth estimation module that generates multi-scale depth priors without computational overhead. Second, we design dynamic depth modulation kernels that exploit these geometric cues for haze-resilient detection. The specific network details are shown in Fig.~\ref{fig:network}.

\textbf{Depth Prior Generation:} We reformulate depth estimation as an auxiliary task through multi-scale depth prediction. Unlike conventional encoder-decoder architectures requiring dedicated upsampling branches \citep{TPAMI2023DetectionFriendlyDehazing}, our DeCoDet directly operates on the backbone's pyramidal features $P = \{P_3, P_4, ..., P_7\}$. At each scale $n$, we apply $M$ depth-specific convolutions:
\begin{equation}
D^{m}=\mathrm{ReLU}(\mathrm{BN}(\mathrm{Conv}_{3\times3}^{m}(D^{m-1}))),\left(m=1,2,\cdots,M\right) ,
\end{equation}
where $D_{n}^{0} = P_{n}$ for each scale $n \in \{3,4,\ldots,7\}$. The final features $D_{n}^{M}$ at each scale are then processed through a $1 \times 1$ convolution to generate the corresponding depth maps $\{D_3,\ldots,D_7\}$, capturing geometric structures across scales. The Mmulti-Scale Depth Prior (MSDP) design effectively handles extreme scale variations in drone imagery while remaining computationally efficient.

\textbf{Depth-Conditioned Feature Modulation:} Building upon the estimated depth priors, we introduce Depth-Conditioned Kernel (DCK) to adaptively modulate detection features. The DCK $\mathcal{K} \in \mathbb{R}^{K\times K\times G}$ at position $(i,j)$ of detection feature $\mathbf{Y}\in\mathbb{R}^{H\times W \times C}$ is generated by:
\begin{equation}
\mathcal{K}_{i,j} = \mathbf{W}_2 \cdot \sigma(\mathbf{W}_1 \cdot D[i,j]).
\end{equation}  

\vspace{-0.5\baselineskip}

 Here, $K$ indicates the kernel size related to spatial modulation range, $G$ is the number of groups sharing a kernel, and $\mathbf{W}_1 \in \mathbb{R}^{C/r \times C}$, $\mathbf{W}_2 \in \mathbb{R}^{(K^2G) \times C/r}$ form a bottleneck structure with reduction ratio $r$ for efficiency. The function $\sigma$ denotes batch normalization and a non-linear activation function, which enhances expressiveness. The modulated feature $\mathbf{Y}'$ is computed via depth-adaptive kernel:
\begin{equation}
\mathbf{Y}_{i,j,g}^{\prime}=\sum_{(u,v)\in\Delta K}\mathcal{K}_{i,j,u+\lfloor K/2\rfloor,v+\lfloor K/2\rfloor,\lceil gG/C \rceil} \odot \mathbf{Y}_{i+u,j+v,g}.
\end{equation}
Here, $\Delta_K = [-\lfloor K/2\rfloor,\ldots,\lfloor K/2\rfloor]\times[-\lfloor K/2\rfloor,\ldots,\lfloor K/2\rfloor] \subset\mathbb{Z}^2$ refers to the set of offsets in the neighborhood when applying the kernel centered at a position ($\times$ indicates Cartesian product here). $g \in \{1,2,\ldots,G\}$ represents the group number.

This architecture establishes three critical properties: 1) Geometric awareness through explicit depth conditioning, 2) Spatial adaptivity via position-specific kernels, and 3) Scale consistency through pyramidal feature alignment.

\subsection{Progressive Domain Fine-Tuning with Scale-Invariant Supervision}

% \textbf{Cross-Domain Curriculum Adaptation:}
Training a robust object detector for real-world hazy drone imagery presents a significant challenge due to the inherent discrepancies between clear, simulated hazy, and real hazy data distributions. While fine-tuning on synthetic data may not generalize well to real-world scenarios, training solely on limited real-world hazy data often leads to overfitting. To bridge the synthetic-to-real domain gap, we propose a Progressive Domain Fine-Tuning (PDFT), as conceptually outlined in Algorithm 1 and visually depicted in Fig.~\ref{fig:dataset and PDFT} (b).

\begin{wrapfigure}{r}{0.49\textwidth}
\label{alg:progressive_finetune} 
\vspace{-15pt}  
\begin{minipage}{0.49\textwidth}
\fontsize{6.5pt}{7.5pt}\selectfont  
\center  
\framebox{%  
\begin{minipage}{0.95\linewidth}
\vspace{2pt}  
\textbf{Algorithm 1:} Progressive Domain Fine-tuning 
\vspace{1pt}  
\hrule  
\vspace{2pt}  
\textbf{Require:} Pre-trained model $M_{\theta_0}$ from source $\mathcal{D}_s$, intermediate domain $\mathcal{D}_m$, target domain $\mathcal{D}_t$, learning rate $\eta_1$, epochs $E_1$, $E_2$, decay $\gamma$ \\
\textbf{Ensure:} Domain-adapted model $M_{\theta_t}$ \\
1: \textbf{Stage 1:} normal → simulated haze (intermediate domain) \\
2: Initialize $\theta_m \leftarrow \theta_0$, freeze $\theta_m^1$ (first backbone stage) \\
3: \textbf{for} epoch $=1$ to $E_1$ \textbf{do} \\
4: \hspace{0.4em} \textbf{for} batch $b_m$ in $\mathcal{D}_m$ \textbf{do} \\
5: \hspace{0.8em} Compute $\mathcal{L}_m = \mathcal{L}_{\textrm{task}}(M_{\theta_m}(b_m))$ \\
6: \hspace{0.8em} Update: $\theta_m \backslash \theta_m^1 \leftarrow \theta_m \backslash \theta_m^1 - \eta_1 \cdot \nabla \mathcal{L}_m$ \\
7: \hspace{0.4em} \textbf{end for} \\
8: \textbf{end for} \\
9: Intermediate domain model: $M_{\theta_m}$ \\
10: \textbf{Stage 2:} simulated → real haze (target domain) \\
11: Initialize $\theta_t \leftarrow \theta_m$, freeze $\{\theta_t^1, \theta_t^{2:k}\}$ (first \& middle stages) \\
12: Set learning rate $\eta_2 = \eta_1 \cdot \gamma$ ($\gamma < 1$) \\
13: \textbf{for} epoch $=1$ to $E_2$ \textbf{do} \\
14: \hspace{0.4em} \textbf{for} batch $b_t$ in $\mathcal{D}_t$ \textbf{do} \\
15: \hspace{0.8em} Compute $\mathcal{L}_t = \mathcal{L}_{\textrm{task}}(M_{\theta_t}(b_t))$ \\
16: \hspace{0.8em} Update: $\theta_t \backslash \{\theta_t^1, \theta_t^{2:k}\} \leftarrow \theta_t \backslash \{\theta_t^1, \theta_t^{2:k}\} - \eta_2 \cdot \nabla \mathcal{L}_t$ \\
17: \hspace{0.4em} \textbf{end for} \\
18: \textbf{end for} \\
19: \textbf{return} $M_{\theta_t}$  
\vspace{1pt}  
\end{minipage}}
\end{minipage}  
\vspace{-25pt}
\end{wrapfigure}

\textit{1) Preliminary Adaptation}: We initialize our model with weights pre-trained on ImageNet and fine-tune it on our synthetic hazy dataset $\mathcal{D}_s$ while freezing only the initial backbone layer. This allows the network to adapt its intermediate-level representations to the characteristics of simulated adverse weather conditions, bridging the gap between the source (ImageNet) and the intermediate (simulated haze) domains.

\textit{2) Progressive Adaptation}: We further fine-tune the model on our real-world hazy drone dataset $\mathcal{D}_t$, this time freezing both the early and middle backbone layers responsible for extracting low and mid-level features. We also employ a reduced learning rate in this stage. This targeted fine-tuning allows the network to adapt its higher-level features to the nuances of real-world haze, effectively transferring knowledge from the simulated to the target (real haze) domain while preserving the learned feature extraction capabilities and mitigating overfitting on the smaller real-world dataset.

\textit{3) Depth-Consistent Robust Learning}: To ensure effective learning of depth information, we employ a loss function that integrates both detection and depth estimation objectives. For stable depth learning, particularly from potentially noisy pseudo-depth maps, we propose the \textbf{Scale Invariant Refurbishment Loss (SIRLoss)}, which utilizes a scale-invariant error metric \citep{NIPS2014scale-invariant} applied to logarithmically transformed depth values. To mitigate the impact of erroneous pseudo-labels, we incorporate a label refurbishment strategy, where the refurbished depth label is computed as a weighted average of the prediction $y$ and a pseudo label $y^{*}$ (obtained through techniques like temporal averaging or ensembling, with a confidence factor $\alpha=0.7$). The SIRLoss is then defined as:
    \begin{equation}
\mathcal{L}_{Dep} = \frac{1}{n}\sum_{i}(\Delta d_i)^2 - \frac{1}{n^2}(\sum_{i}\Delta d_i)^2
\end{equation} 
where $\Delta d_i = \log(y_i) - \log(\alpha y^*_i + (1-\alpha){y}_i)$, with $\alpha$ controlling pseudo-label confidence.

%% file: arXiv-contents/experiment.tex
\vspace{-0.5em}
\section{Experiments} \label{sec:experiment}
\vspace{-0.5em}

\subsection{Implementation Details} \label{subsec:setting}

Our DeCoDet architecture adopts VFNet~\citep{CVPR2021VFNet} with an ImageNet-pretrained ResNet-50 backbone as the baseline. During training, we apply random horizontal flipping (50\% probability) and normalize inputs to a resolution of $1333\times800$. The network is optimized via Stochastic Gradient Descent (SGD) over 12 epochs, beginning with an initial learning rate of 0.02 (a constant warmup followed by $10\times$ reductions at epochs 8 and 11), a batch size of 2, a momentum of 0.938, and a weight decay of 0.0001. All experiments are conducted using PyTorch 2.0 on NVIDIA RTX 3090 GPUs.

\input{tables/det_bench}

In Algorithm 1, $\gamma$ is specified as 0.1, and $k$ is set to 3. The depth maps for DeCoDet are generated as described in Sec.~\ref{sec:dataset} and have been anonymously released to facilitate the reproduction of results. Detection accuracy is measured via mean Average Precision (mAP) and Average Precision (AP), while efficiency is gauged by Giga Floating-Point Operations (GFLOPs), model parameters, and frames per second (FPS). Additionally, to evaluate the dehazing method's performance, we employ two widely recognized image restoration metrics: Peak Signal-to-Noise Ratio (PSNR) and Structural Similarity Index Measure (SSIM).

\vspace{-0.5em}
\subsection{Benchmark} \label{subsec:sota} 
\vspace{-0.5em}

\noindent\textbf{Detector Benchmark}
As shown in Tab.~\ref{tab:DetBench}, we benchmark 19 leading object detection algorithms for foggy drone detection using our HazyDet dataset. Fog-specialized detectors~\citep{AAAI2022IAYOLO, ComputerGraphicsForum2022TogetherNet, ICIP2021msyolo} maintain their original protocols, while all other models employ our progressive fine-tune approach from Algorithm 1, which involves training intermediate models on simulated data before real-world fine-tuning. We ensure fair comparison through consistent data augmentation protocols, with exceptions only for methods requiring specific augmentation techniques (e.g., DAB-DETR \citep{ICLR2022DABDETR} and Deformable DETR \citep{ICLR2021DeformableDETR}).

\vspace{-0.5em}
\input{tables/ablation_all}

Consistent performance trends between simulated and real-world environments validate our dataset. Our DeCoDet method outperforms all competitors with model achieving 52.0\% mAP on simulated data and 38.7\% on real-world data, while maintaining higher speed (59.7 FPS) and efficiency (34.62M parameters), making it ideal for deployment in challenging foggy environments.

\input{tables/dehaze_bench}

\noindent\textbf{Dehazing Benchmark}
A systematic evaluation of several dehazing models is conducted on HazyDet. Faster R-CNN~\citep{TPAMI2017FasterRCNN} is adopted as the baseline detector, which is trained exclusively on clear drone images. During testing, hazy images are first dehazed before being fed into the detector, and image quality metrics, including SSIM and PSNR, are computed with respect to the corresponding clear reference images. 

As shown in Tab.~\ref{tab:DefogBench}, most dehazing models yield only marginal improvements in image quality and frequently fail to enhance, or may even diminish, detection accuracy, especially on real-world data. This limitation may be attributed to insufficient adaptation to drone perspectives. Only a few models, such as RIDCP~\citep{CVPR2023RIDCP}, achieve significant gains in detection performance. Furthermore, higher PSNR and SSIM scores do not necessarily translate to improved detection accuracy, highlighting the complex relationship between restoration quality and detection performance. Even the best-performing dehazing models are unable to match the accuracy of detectors trained directly on hazy images, emphasizing the importance of domain-specific training for UAV detection in hazy environments. Qualitative dehazing results from different models are also visualized in Fig.~\ref{fig:dehazing_result}.

\subsection{Ablation Study} \label{subsec:ablation}

Tab.~\ref{tab:ablation} presents our ablation studies validating DeCoDet’s components and parameter settings. Progressive fine-tuning leads to a substantial boost in real-world performance. Introducing multi-scale depth priors gives VFNet~\citep{CVPR2021VFNet} moderate improvements ($+0.3\%$ mAP on the test set, $+0.2\%$ on real data). The combination of DCK and MSDP produces significant gains ($+0.8\%$ test, $+1.1\%$ real), while SIRLoss offers only limited benefits ($+0.2\%$ test, $+0.1\%$ real). The complete model, integrating all components, achieves the best results ($52.0\%$ mAP on the test set, $38.0\%$ on real data), indicating their complementary effects.

Parameter optimization shows that using three depth-specific convolutional layers ($M=3$) in the detection head balances performance and computational cost. DCK performs best with a $7\times7$ kernel and $16$ groups, capturing sufficient contextual information without introducing noise. In contrast, smaller kernels ($3\times3$) lead to large drops in accuracy ($-9.1\%$ test, $-6.8\%$ real), and excessive grouping brings only marginal gains at higher cost. The optimal SIRLoss weight is $\beta = 0.2$; setting $\beta$ lower ($0.05$) causes underfitting ($-3.8\%$ test, $-3.1\%$ real), and higher ($1.0$) reduces accuracy ($-1.9\%$ test, $-1.1\%$ real).

\vspace{-0.5em} 
\begin{figure*}[t]
    \centering
    \includegraphics[width=0.95\linewidth]{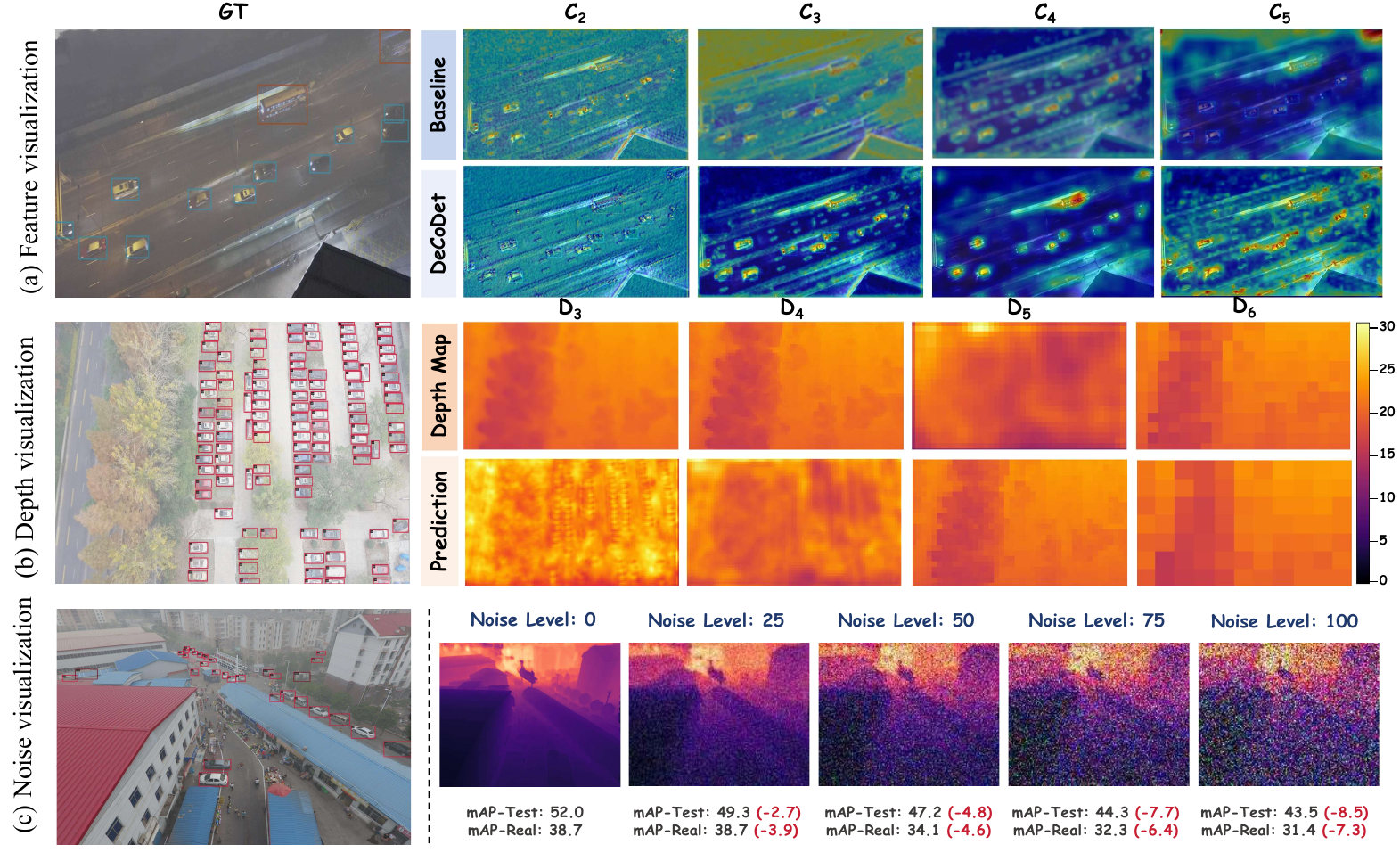}
    % \captionsetup{font={scriptsize}}
    \caption{\textbf{Network visualization and robustness analysis.} (a) Feature visualization comparing baseline and DeCoDet models under foggy conditions, displaying ground truth (left) alongside Grad-CAM \citep{ICCV2017GradCAM} activation maps across backbone layers C2-C5. (b) Depth estimation visualization across network layers, showing original ground truth (left) and corresponding depth maps at different level(right). The upper row presents learned depth maps while the lower row shows network predictions. Color bar indicates depth in meters. (c) Network robustness verification. The left columns show the original ground truth, while the right columns present the depth maps used for DeCoDet training. The noise level denotes the variance of the added Gaussian noise, with 0 indicating no noise. Values at the bottom reflect the network’s performance with noise-aware labels on both simulated and real-world data.}
    \label{fig:merged_visual}
    \vspace{-1.5em} 
\end{figure*}

\subsection{Visualization and Robustness Analysis} \label{subsec:visual}

\noindent\textbf{Feature visualization.}
Fig.~\ref{fig:merged_visual}~(a) compares backbone feature heatmaps between baseline and DeCoDet under foggy conditions. DeCoDet precisely identifies potential target regions, focusing network attention more effectively. This targeted focus improves detection accuracy in challenging foggy environments where traditional models struggle. By incorporating depth information, DeCoDet enhances feature discrimination while reducing haze interference, demonstrating robust performance in adverse weather.

\noindent\textbf{Depth-Cue Visualization.}
Fig.~\ref{fig:merged_visual}~(b) reveals the depth information learned by DeCoDet. Conventional depth estimation yields smooth outputs but misses small targets. DeCoDet aligns depth globally at high levels while emphasizing detection cues locally, boosting sensitivity to tiny, haze-obscured objects and improving detection in low visibility.

\noindent\textbf{Robustness Verification.}
We assess DeCoDet’s robustness by adding varying levels of gaussian noise to training depth maps. As illustrated in Fig.~\ref{fig:merged_visual}~(c), network performance declines with increasing noise, especially at higher levels. These results highlight the importance of high-quality depth labels for learning accurate depth cues and stable training, suggesting that enhancing depth estimation accuracy is vital for future performance improvements.

%% file: tables/det_bench.tex
\begin{table}[t] 
% \captionsetup{font={scriptsize}}  
\caption{\textbf{Comparison of the performance of different SOTA detectors on the HazyDet dataset.} Bold indicates the highest performance, and underline highlights the second highest. Rankings are across all models.}  
\vspace{-0.5em}
\label{tab:DetBench} 
\renewcommand{\arraystretch}{0.9}
\begin{center}  
\begin{adjustbox}{max width=\textwidth}  
\begin{tabular}{llccccccccccccc}  
\toprule  
\multirow{2}{*}{Type} & \multirow{2}{*}{Model} & \multirow{2}{*}{Epoch} & \multirow{2}{*}{FPS} & \multirow{2}{*}{Para (M)} & \multirow{2}{*}{GFLOPs} & \multicolumn{4}{c}{AP on Synthetic Test-Set} & & \multicolumn{4}{c}{AP on Real-World Test-Set} \\  
\cmidrule(lr){7-10} \cmidrule(lr){12-15}  
                       &                       &                        &                        &                           &                          & mAP & Car & Truck & Bus  &  & mAP & Car  & Truck & Bus       \\  
\midrule  
\multirow{3}{*}{Specified}  
                       & IAYOLO \citep{AAAI2022IAYOLO}       & 273  & 22.7 & 61.80  & 37.30        & 38.3 & 44.1 & 22.2 & 48.6 &  & 22.4 & 41.9 & 8.0 & 17.3  \\  
                       & TogetherNet \citep{ComputerGraphicsForum2022TogetherNet} & 300  & 49.5 & 69.2  & 19.71         & 44.6 & 53.4 & 25.4 & 55.0 &  & 25.2 & 48.2 & 11.3 & 16.15 \\ 
                       & MS-DAYOLO \citep{ICIP2021msyolo} & 300  & 57.9 & 40.0 & 44.20                             & 48.3 & 59.4 & 28.5 & 57.0 &  &36.5 &51.1 & 21.0 & 37.4 \\               
\midrule          
\multirow{8}{*}{One-Stage}  
                       & YOLOv3 \citep{Arxiv2018yolov3}        & 273 & 68.1 & 61.63 & 20.19   & 35.0 & 36.1 & 21.4 & 47.5 &  & 30.7 &52.3 & 12.2 & 27.5 \\  
                       & GFL \citep{CVPR2021GFLv2}             & 12  & 51.1 & 32.26 & 198.65  & 36.8 &50.3 & 11.5 & 48.5 &  &32.5 &50.9 & 15.6 & 31.0 \\  
                       & YOLOX \citep{Arxiv2021yolox}          & 300 & 71.2 & 8.94  & 13.32   & 42.3 &53.1 & 23.0 & 51.2 &  &35.4 &52.9 & 19.9 & 33.5 \\  
                       & FCOS \citep{tianFCOSFullyConvolutional2019} & 12 & 60.6 & 32.11 & 191.48  & 45.9 &54.4 & 27.1 & 56.2 &  &32.7 &50.7 & 16.0 & 31.4 \\  
                       & VFNet \citep{CVPR2021VFNet}           & 12  & 59.9 & 32.71 & 184.32  & 49.5 &58.2 & 30.8 & 59.4 &  &35.6 &52.5 & 18.4 & 36.0 \\  
                       & ATTS \citep{CVPR2020ATTS}             & 12  & 49.6 & 32.12 & 195.58  & 50.4 &58.5 & 32.2 & 60.4 &  &36.4 &52.4 & 19.5 & 37.2 \\  
                       & DDOD \citep{ACMICMM2021DDOD}          & 12  & 47.6 & 32.20 & 173.05  & 50.7 &\underline{59.5} & 32.1 & 60.4 &  &37.1 &51.8 & 21.0 & \underline{38.5} \\  
                       & TOOD \citep{ICCV2021TOOD}             & 12  & 48.0 & 32.02 & 192.51  & 51.4 &58.4 & 33.6 & \underline{62.2} &  &36.7 &52.5 & 20.0 & 37.6 \\  
\midrule  
\multirow{5}{*}{Two-Stage}  
                       & Sparse RCNN \citep{TPAMI2023SparseRCNN} & 12  & 47.1 & 108.54 & 147.45 & 27.7 &33.0 & 14.2 & 35.6 &  &20.8 &31.0 & 10.4 &21.0 \\  
                       & Faster RCNN \citep{TPAMI2017FasterRCNN} & 12  & 46.4 & 41.35  & 201.72 & 48.7 &56.3 & 30.5 &59.3 &  &33.4 &54.2 &12.5 &33.5 \\  
                       & Libra RCNN \citep{CVPR2019LibraRCNN} & 12  & 44.7 & 41.62  & 209.92 & 49.0 &57.3 &30.4 &59.3 &  &34.5 &54.5 &14.5 &34.6 \\  
                       & Grid RCNN \citep{CVPR2019GridRCNN}  & 12  & 47.2 & 64.46  & 317.44 &50.5 &58.1 &32.8 &50.7 &  &35.2 &54.3 &18.6 &32.8 \\  
                       & Cascade RCNN \citep{CVPR2018CascadeRCNN} & 12  & 42.9 & 69.15  & 230.40 &\underline{51.6} &59.0 &\textbf{34.2} &\underline{61.7} &  &\underline{37.2} &\underline{54.8} &19.0 &37.9 \\  
\midrule  
\multirow{3}{*}{End2End} 
                      & Conditional DETR \citep{ICCV2019ConditionDETR}       & 50  & 40.3 & 43.55  & 94.17   &30.5 &42.1 &12.6 &36.8 &  &25.8 &46.0 &10.1 &21.3 \\
                       & DAB DETR \citep{ICLR2022DABDETR}       & 50  & 41.7 & 43.70  & 97.02   &31.3 &36.8 &15.1 &42.3 &  &27.2 &47.4 &12.2 &22.0 \\  
                       & Deform DETR \citep{ICLR2021DeformableDETR} & 50  & 50.9 & 40.01  & 203.11 &51.5 &58.4 &33.9 &\textbf{62.3} &  &36.9 &52.1 &\underline{21.1} &37.5 \\  
\midrule  
    \rowcolor[rgb]{0.9,0.9,0.9} & $\star$ \textbf{DeCoDet (Ours)}                           & 12 & 59.7 & 34.62  & 225.37 &\textbf{52.0} &\textbf{60.5} &\underline{34.0} &61.9 &  &\textbf{38.7} &\textbf{55.0} &\textbf{21.9} &\textbf{39.2} \\  
\bottomrule  
\end{tabular}  
\end{adjustbox}  
\end{center}  
\vspace{-1.5em}  
\end{table}

%% file: tables/ablation_all.tex
\begin{table}[t] 
\caption{\textbf{Ablation study results.}  
The baseline is VFNet. Progressive: progressive fine-tune; MSDP: multi-scale depth prior generation; DCK: depth-conditioned kernel module; SIRLoss: scale invariant refurbishment loss; M: number of depth-specific convolutions; Kernel: DCK kernel size; Group: DCK group size; $\beta$: SIRLoss weight. mAP--Test and mAP--Real denote average precision on synthetic Test-set and real-world Test-set. Bold indicates the highest performance.}
\vspace{-0.5em}
\label{tab:ablation}  
\renewcommand{\arraystretch}{0.85} % 缩小行间距
\begin{adjustbox}{max width=\textwidth}  
\begin{tabular}{@{}lcccccccccccc@{}}  
\toprule  
Method 
& Progressive 
& MSDP   
& DCK   
& SIRLoss   
& M  
& Kernel  
& Group  
& weight $\beta$  
& Para(M)   
& GFLOPs   
& mAP--Test   
& mAP--Real \\
\midrule 
\multirow{6}{*}{DeCoDet}
& - & - & - & - & - & - & - & -   
& 32.71 & 184.32 & 49.5 & 25.6 \\
& \checkmark & - & - & - & - & - & - & -   
& 32.71 & 184.32 & 49.5 & 35.6 \\
& \checkmark & \checkmark & - & - & 3 & - & - & -   
& 34.59 & 221.14 & 49.8 & 35.8 \\
& \checkmark & \checkmark & \checkmark & - & 3 & 7 & 16 & -   
& 34.61 & 230.41 & 50.6 & 36.9 \\
& \checkmark & \checkmark & - & \checkmark & 3 & - & - & 0.2   
& 34.59 & 221.14 & 50.0 & 35.9 \\
& \checkmark & \checkmark & \checkmark & \checkmark & 3 & 7 & 16 & 0.2   
& 34.61 & 230.41 & \textbf{52.0} & \textbf{38.0} \\
\midrule  
\multirow{12}{*}{\makecell[l]{Ablation\\Study}}  
& \multirow{12}{*}{\checkmark}  
& \multirow{12}{*}{\checkmark} 
& \multirow{12}{*}{\checkmark}  
& \multirow{12}{*}{\checkmark}  
& 1 & 7 & 16 & 0.2  
& 31.39 & 156.68 & 50.1 & 35.6 \\
&&&& & 2 & 7 & 16 & 0.2  
& 33.16 & 202.75 & 50.6 & 36.0 \\
&&&& & 3 & 3 & 16 & 0.2  
& 34.53 & 224.26 & 42.9 & 31.2 \\
&&&& & 3 & 5 & 16 & 0.2  
& 34.57 & 226.31 & 49.6 & 35.9 \\
&&&& & 3 & 9 & 16 & 0.2  
& 34.68 & 237.38 & 51.4 & 37.3 \\
&&&& & 3 & 7 & 1 & 0.2  
& 34.52 & 224.46 & 49.2 & 34.3 \\
&&&& & 3 & 7 & 4 & 0.2  
& 35.54 & 226.31 & 51.0 & 36.8 \\
&&&& & 3 & 7 & 64 & 0.2  
& 34.92 & 249.86 & 51.6 & 37.2 \\
&&&& & 3 & 7 & 16 & 0.05  
& 34.61 & 230.41 & 48.2 & 34.9 \\
&&&& & 3 & 7 & 16 & 0.1  
& 34.61 & 230.41 & 50.3 & 36.8 \\
&&&& & {\cellcolor{gray!15}}3 & {\cellcolor{gray!15}}7 & {\cellcolor{gray!15}}16 & {\cellcolor{gray!15}}0.2  
& {\cellcolor{gray!15}}34.61 & {\cellcolor{gray!15}}230.41 & {\cellcolor{gray!15}}\textbf{52.0} & {\cellcolor{gray!15}}\textbf{38.0} \\
&&&& & 3 & 7 & 16 & 1.0  
& 34.61 & 230.41 & 50.1 & 36.9 \\
\bottomrule  
\end{tabular}  
\end{adjustbox}  
\vspace{-1.5em}  
\end{table}

%% file: tables/dehaze_bench.tex
\begin{wraptable}{r}{0.6\textwidth}  
\vspace{-1.2em}  
\centering  
% \captionsetup{font={scriptsize}}  
\caption{\textbf{Performance comparison of SOTA dehazing methods on HazyDet.} mAP--Test and mAP--Real denote average precision on synthetic Test-set and real-world Test-set. 
% ``-'' means missing; bold is best, underline is second best.
}  
\label{tab:DefogBench}  
\scriptsize  
\setlength{\tabcolsep}{3pt}  % 更紧凑的列间距  
\renewcommand{\arraystretch}{1.05}  
\begin{tabular}{lcccc}  
\toprule  
Method & PSNR$\uparrow$ & SSIM$\uparrow$ & mAP-Test & mAP-Real \\
\midrule  
Faster RCNN~\citep{TPAMI2017FasterRCNN}            & -         & -         & 39.5              & 21.5              \\
GridDehaze~\citep{ICCV2019GridDehazeNet}  & 12.66     & 0.713     & 38.9 {\scriptsize\textcolor[rgb]{1,0.4,0.471}{(-0.6)}}  & 19.6 {\scriptsize\textcolor[rgb]{1,0.4,0.471}{(-1.9)}}  \\
MixDehazeNet~\citep{Arxiv2023mixdehazenetmixstructure}  & 15.52     & 0.743     & 39.9 {\scriptsize\textcolor[rgb]{0.125,0.761,0.596}{(+0.4)}}   & 21.2 {\scriptsize\textcolor[rgb]{1,0.4,0.471}{(-0.3)}}   \\
DSANet~\citep{NeuralNetworks2024DSANet}  & 19.01     & 0.751     & 40.8 {\scriptsize\textcolor[rgb]{0.125,0.761,0.596}{(+1.3)}}   & 22.4 {\scriptsize\textcolor[rgb]{0.125,0.761,0.596}{(+0.9)}}  \\
FFA~\citep{AAAI2020FFANet}  & 19.25     & 0.798     & 41.2 {\scriptsize\textcolor[rgb]{0.125,0.761,0.596}{(+1.7)}}   & 22.0 {\scriptsize\textcolor[rgb]{0.125,0.761,0.596}{(+0.5)}}  \\
DehazeFormer~\citep{TIP2023RSHaze}  & 17.53     & 0.802     & 42.5 {\scriptsize\textcolor[rgb]{0.125,0.761,0.596}{(+3.0)}}   & 21.9 {\scriptsize\textcolor[rgb]{0.125,0.761,0.596}{(+0.4)}}  \\
gUNet~\citep{Arxiv2022gUNet}  & \underline{19.49}   & 0.822     & 42.7 {\scriptsize\textcolor[rgb]{0.125,0.761,0.596}{(+3.2)}}   & 22.2 {\scriptsize\textcolor[rgb]{0.125,0.761,0.596}{(+0.7)}}  \\
C2PNet~\citep{CVPR2023C2PNet}  & \textbf{21.31}  & \textbf{0.832}  & 42.9 {\scriptsize\textcolor[rgb]{0.125,0.761,0.596}{(+3.4)}}   & 22.4 {\scriptsize\textcolor[rgb]{0.125,0.761,0.596}{(+0.9)}}  \\
DCP~\citep{CVPR2009DCP}  & 16.98     & \underline{0.824}   & \underline{44.0} {\scriptsize\textcolor[rgb]{0.125,0.761,0.596}{(+4.5)}}   & \underline{20.6} {\scriptsize\textcolor[rgb]{1,0.4,0.471}{(-0.9)}}   \\
RIDCP~\citep{CVPR2023RIDCP}  & 16.15     & 0.718     & \textbf{44.8} {\scriptsize\textcolor[rgb]{0.125,0.761,0.596}{(+5.3)}}  & \textbf{24.2} {\scriptsize\textcolor[rgb]{0.125,0.761,0.596}{(+2.7)}}  \\
\bottomrule  
\end{tabular}  
\vspace{-1.5em}  
\end{wraptable}

%% file: arXiv-contents/conclusion.tex
\section{Conclusion} \label{sec:conclusion}

In this work, we addressed the critical challenge of drone-view object detection in hazy environments by introducing HazyDet, the first large-scale benchmark of its kind, amalgamating extensive real-world and high-fidelity synthetic hazy imagery. We further proposed the DeCoDet framework, a novel detector that innovatively leverages depth cues via a Depth-Conditioned Kernel, thereby eschewing explicit dehazing. DeCoDet's robustness is significantly enhanced through a Progressive Domain Fine-Tuning strategy and a Scale-Invariant Refurbishment Loss, enabling effective learning across domain shifts and from imperfect depth supervision. Our comprehensive experiments demonstrate that this unified framework establishes a new state-of-the-art on HazyDet, achieving a notable +1.5\% mAP improvement over prior methods on challenging real-world hazy data. 

\section*{Data Availability Statement}
We confirm that data supporting the results of this study can be obtained from \citep{TIP2019RESIDE}, \citep{CVPR2021_4KDehaze} and \href{https://github.com/GrokCV/HazyDet}{HazyDet}.

%% file: appendix.tex
\newpage
\renewcommand{\thetable}{S\arabic{table}}
\renewcommand{\thefigure}{S\arabic{figure}}
\renewcommand{\theequation}{S\arabic{equation}}
\setcounter{equation}{0}
\setcounter{table}{0}
\setcounter{figure}{0}
\appendix
\section*{Appendix} % 将“Appendix”部分放在新页面中间

The appendix presents a detailed analysis of the dataset used, provides comprehensive ablation studies and visualizations, and concludes with a discussion of broader implications and safeguards.

\section{Dataset Details} \label{sec:dataset_analysis}

\subsection{Dataset Construction Details}

\begin{figure}[th!]
    \vspace{-1em}
    \centering
    \includegraphics[width=0.9\linewidth]{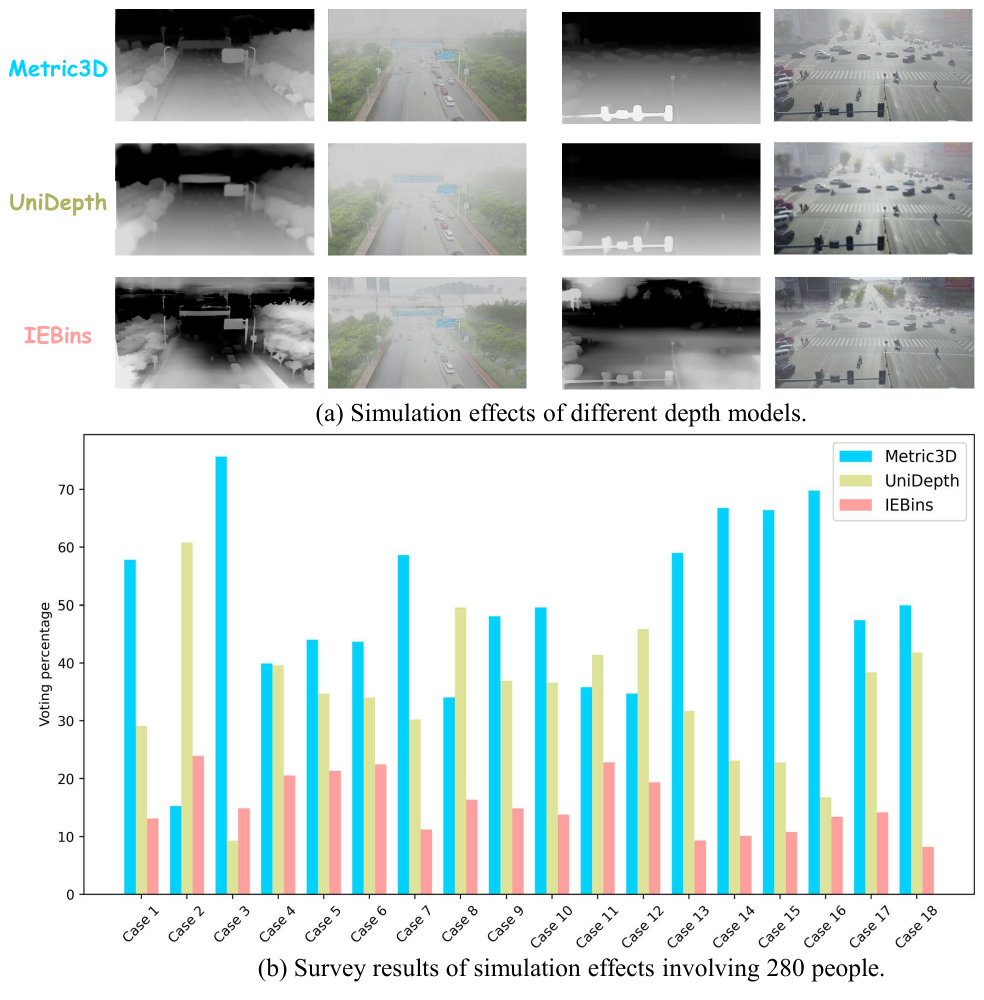}
    \vspace{-0.5em}
    \caption{\textbf{Haze simulation effect comparison.} (a) Subjective visual comparisons among outputs from various depth estimation models under identical ASM parameters. (b) A large-scale questionnaire-based assessment, where the horizontal axis enumerates different test cases and the vertical axis shows the percentage of participant votes.}
    \label{fig:vote}
     \vspace{-0.5em}
\end{figure}

Following the approach in \citep{TIP2019RESIDE,CVPR2021_4KDehaze}, we obtain the scene depth required for the simulation through a depth estimation model. However, the models used in the aforementioned papers exhibit poor generalization ability under unseen drone perspectives. After evaluating state-of-the-art depth estimation methods, we select three models that demonstrate excellent zero-shot learning performance and better generalize to unknown domains \citep{Arxiv2024metric3d,CVPR2024Unidepth,NEURIPS2023IEBins}. The simulation results are shown in Fig.~\ref{fig:vote}~(a). To evaluate the effect of fog simulation under no-reference metrics, we conduct a questionnaire survey of 280 university students and experts to assess the realism, brightness distribution, and fog consistency of the synthesized images across 18 scenes. Fig.~\ref{fig:vote}~(b) shows that the images generated using \citep{Arxiv2024metric3d} closely resemble real-world scenes.

\subsection{Dataset Authenticity.}\label{subsec:authentic}

\begin{wrapfigure}{r}{0.49\textwidth}
    \vspace{-35pt}
    \centering
    % \captionsetup{font={scriptsize}}
    \includegraphics[width=0.5\textwidth]{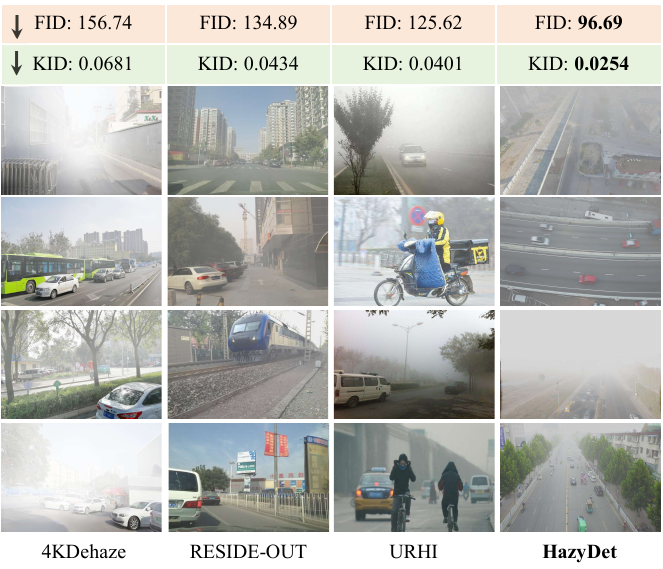}
    \caption{\textbf{Similarity comparison between datasets and real drone-perspective hazy scenes.} The upper section presents objective metrics ($\downarrow$ indicates lower values are better), while the lower section displays sample visualizations from each dataset. The metric calculation is conducted between the mentioned datasets and the real-world data collected in HazyDet.}
    \label{fig:authenticity}
    \vspace{-20pt}
\end{wrapfigure}

To evaluate the authenticity of our synthetic data objectively, we use the Fréchet Inception Distance (FID) \citep{arxiv2021FID} and Kernel Inception Distance (KID) \citep{ICLR2018KID} to test the similarity between the synthetic hazy data and the real data distribution. Fig.~\ref{fig:authenticity} reveals that HazyDet offers closer approximations to real drone-captured foggy conditions than datasets like RESIDE-Out \citep{TIP2019RESIDE} and 4KDehaze \citep{CVPR2021_4KDehaze}. Even compared to the real data within URHI \citep{TIP2019RESIDE} dataset, our approach excels due to its alignment with drone perspectives. While FID and KID indices offer some insight into the quality of synthesized haze images, they share limitations with other blind quality assessment methods, as predicted scores may not always align with human perception. To address this, we conduct a subjective visual comparison of our dataset with existing mainstream datasets, illustrated in Fig.~\ref{fig:authenticity}. The results clearly indicate that HazyDet more accurately mirrors real foggy conditions across different haze levels.

\subsection{Statistics and Characteristics of Instances}

\begin{figure}[thp]
    \centering
    \includegraphics[width=0.95\linewidth]{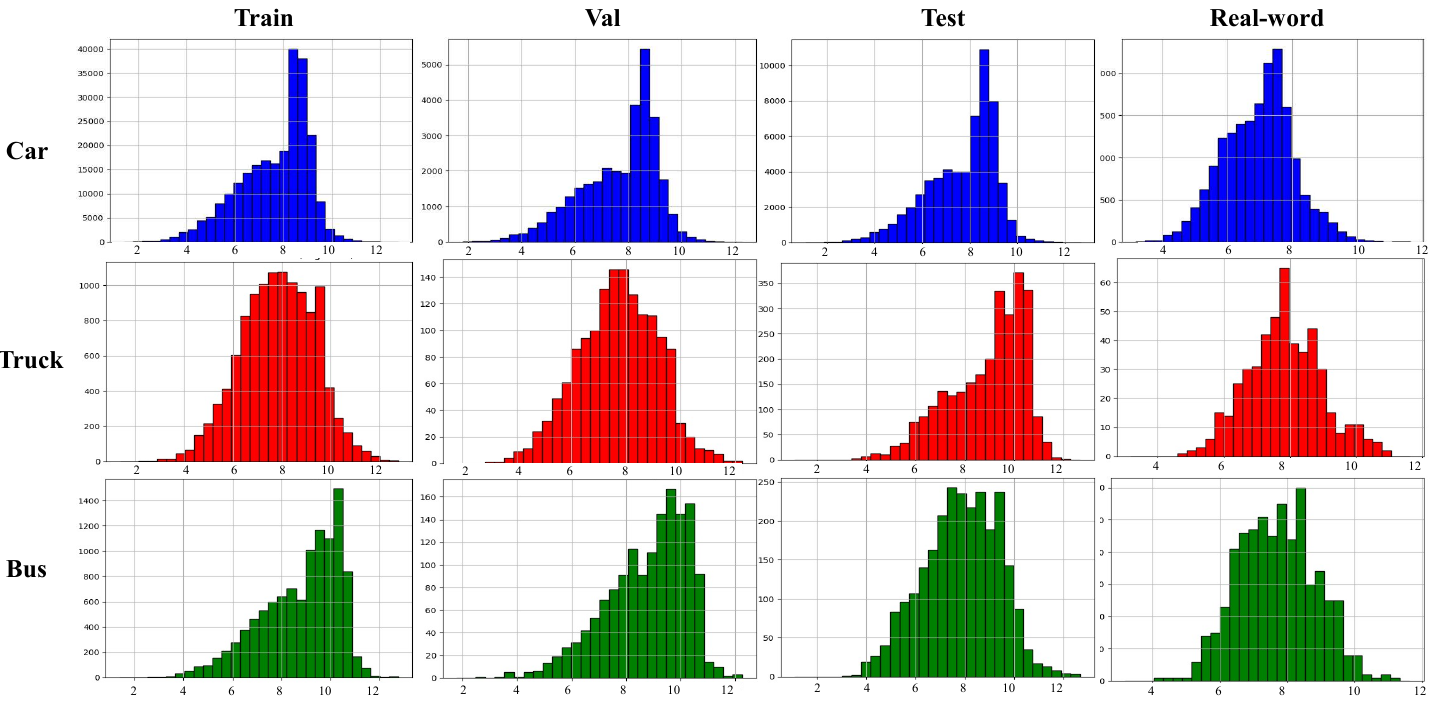}
     % \captionsetup{font={footnotesize}}
    \caption{\textbf{Statistical distribution of instacnes at different scales in HazyDet}. In the bar chart, the horizontal axis represents the logarithmic value (base 2) of the number of pixels occupied by that category of target in the subset, and the vertical axis represents the frequency count.}
    \label{fig:data_scale}
     \vspace{-0.5em}
\end{figure}

Tab.~\ref{tab:dataset_count} provides a systematic presentation of image distribution and instance statistics across subsets in the HazyDet dataset. Comprising 11,000 synthesized images with 365,000 annotated instances, the dataset is partitioned into training, validation, and testing subsets in a stratified 8:1:2 ratio, encompassing multiple object categories including Car, Truck, and Bus. To complete dataset, we have incorporated 600 annotated real-world foggy scene images adhering to the same annotation protocols as synthetic data. This synthetic-real bimodal data framework with high object density establishes an ideal benchmark for robust evaluation of detection models. Fig.~\ref{fig:data_scale} illustrates the distribution of object scales within the dataset. While the distribution remains relatively consistent across different subsets of simulated data, notable differences exist between simulated and real datasets due to image resolution disparities. Regarding object categories, cars and trucks exhibit a distribution skewed toward smaller scales compared to buses. The majority of objects have scale measurements concentrated between $2^6$ (64 pixels) and $2^10$ (1024 pixels), indicating that small objects constitute the predominant portion of instances throughout the dataset.

\subsection{Depth-related Characteristics.}

\begin{figure}
    \vspace{-2em}
    \centering
    \includegraphics[width=0.95\linewidth]{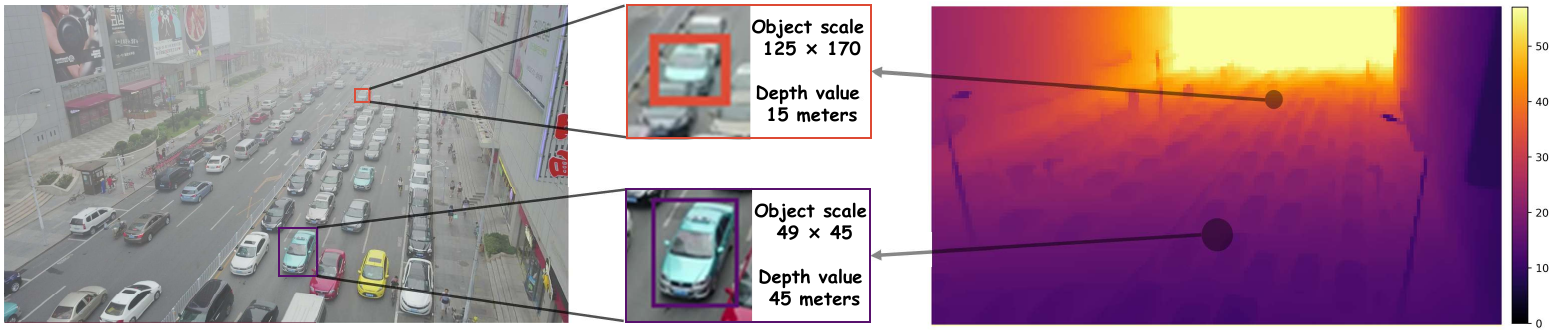}
     % \captionsetup{font={footnotesize}}
    \caption{ \textbf{Depth correlation visualization in HazyDet.} The left figure shows that haze affects images to varying degrees at different depths. At the same time, the same object exhibits significant size differences at different depths, which is related to the depth of the scene in which the object is located.
    }
    \label{fig:depth_ana}
     \vspace{-1em}
\end{figure}

Drone object detection in adverse weather conditions has broader associations with depth information. ASM (Equation \ref{eq1}) shows that under the same atmospheric parameters, there is an exponential correlation between pixel degradation intensity and scene depth cues, indicating that depth maps are closely related to foggy scene distribution. Additionally, the unique flight altitude and shooting angle of drones enhance the perspective effect in images, highlighting the clear relationship between target size and depth. Targets at closer distances appear larger, as shown in Fig.~\ref{fig:depth_ana} (a), which is consistent with intuitive expectations. Objective statistics on data in HazyDet also demonstrate this, as shown in Fig.~\ref{fig:depth_ana} (b). These depth-related insights provide important background information for interpreting drone images in foggy environments and are expected to support various scene interpretation tasks.

\section{Additional Experiments}\label{sec.addexp}

\subsection{Comparative Experiments}\label{subsec:compare_exp}

\vspace{-0.5em}

\input{tables/decodet_compare}

\vspace{-1.5em}

We analyze how different experimental setups in DeCoDet affect performance to better understand the role and mechanisms of depth cues.

\noindent\textbf{Effectiveness of Depth Loss.}
Tab.~\ref{tab:depth_and_noise} outlines the performance impacts of various depth estimate loss functions. Traditional loss functions like SmoothL1 and MSE focus on absolute differences, making them susceptible to noise in pseudo-labels and thus limiting effective depth-cue utilization for condition. In contrast, SIRLoss maintains scale invariance and enhances label refurbishment, yielding superior mAP scores.

\begin{figure*}[th!]
    \centering
    \vspace{-1em}
    \includegraphics[width=0.95\linewidth]{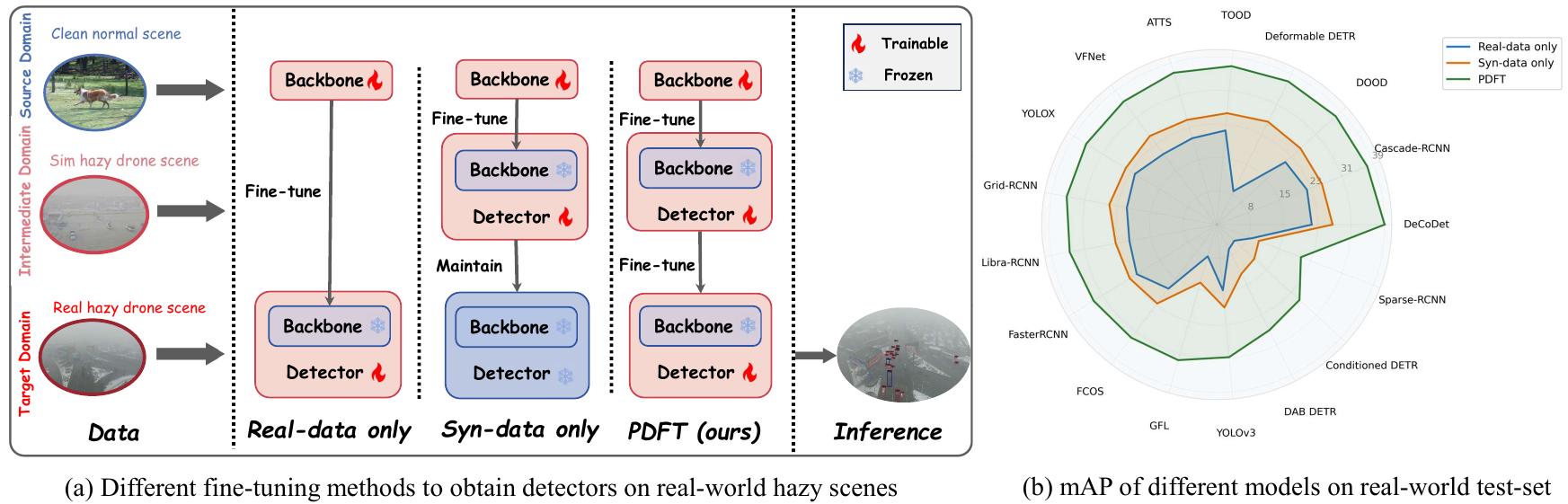}
     % \captionsetup{font={footnotesize}}
    \caption{\textbf{Performance comparison of different fine-tuning methods in real-world drone hazy scenarios.} (a) We compare three different methods: the first fine-tunes only on the real dataset, the second fine-tunes only on simulated data, and the third method (PDFT) uses simulated data as a bridge to implement progressive fine-tuning. (b) Performance comparison of different fine-tuning methods shows that PDFT improves the detection performance of all model groups in real UAV hazy scenarios.}
    \label{fig:fine-tune_compare}
     \vspace{-1.5em}
\end{figure*}

\noindent\textbf{Effectiveness of Depth Map.}
We conduct experiments to evaluate the impact of depth maps generated by various estimation models on DeCoDet's performance. Beyond the models discussed in Sec.~\ref{sec:dataset}, we include VA-DepthNet \citep{ICLR2023vadepthnet} and ZoeDepth \citep{Arxiv2023zoedepth}. Tab.~\ref{tab:depth_and_noise} demonstrates that prediction of Metric3D \citep{Arxiv2024metric3d} achieved superior results, attributed to its exceptional depth estimation accuracy and generalization capacity in novel environments. These results underscore the critical necessity of accurate depth maps for enhancing detection capabilities.

\noindent\textbf{Effectiveness of PDFT.}
As illustrated in Fig.~\ref{fig:fine-tune_compare}, we conduct experiments to evaluate the impact of different fine-tuning strategies on model performance in real-world hazy scenarios. Direct fine-tuning on real-world data, however, consistently encounters severe domain shift and overfitting due to significant semantic and scale discrepancies, resulting in suboptimal performance across all models. In contrast, fine-tuning with simulated data leads to notable improvements, indicating that simulated datasets provide beneficial semantic features for drone-view detection under hazy conditions and sufficient scale to mitigate overfitting. Building on this observation, our proposed PDFT method utilizes simulated data as a bridging domain, effectively alleviating these challenges. As a result, our approach consistently delivers performance gains across all evaluated models.

\subsection{Detection Result Visualizations}\label{subsec:false}

\begin{figure*}[ht!]
    \centering
    \includegraphics[width=0.95\linewidth]{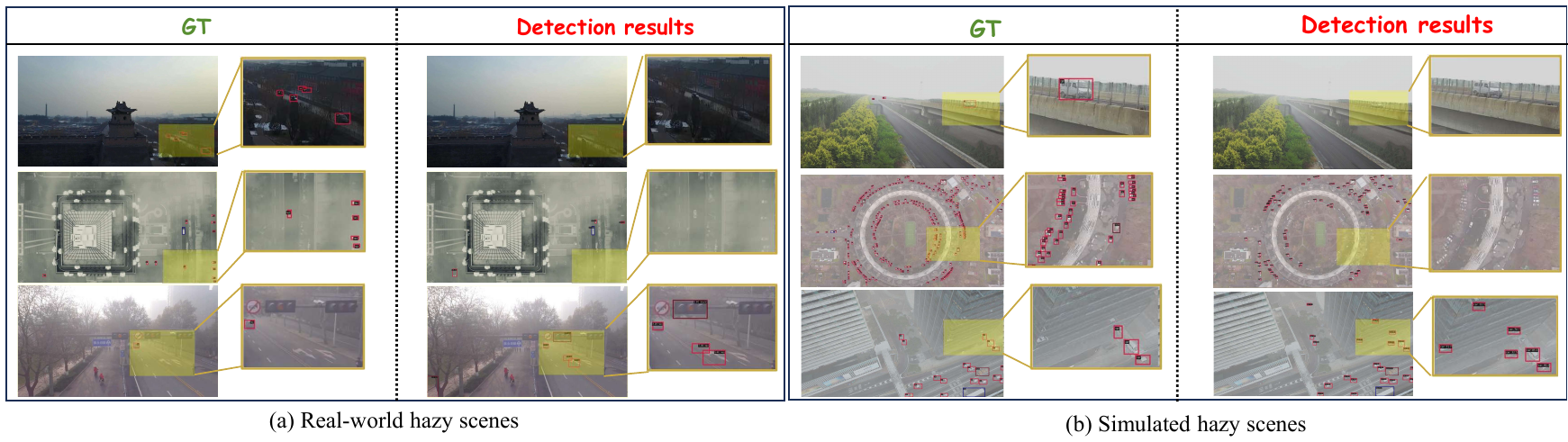}
     % \captionsetup{font={footnotesize}}
    \caption{\textbf{Challenging detection scenarios for DeCoDet in foggy drone imagery.} (a) Missed detections due to background occlusion despite depth information. (b) Detection failures of small-scale targets from elevated drone perspectives in hazy conditions. (c) False positives triggered by fog-induced feature distortion.}
    \label{fig:false_samples}
     \vspace{-0.5em}
\end{figure*}

Fig.~\ref{fig:false_samples} presents the visualization of our network detection results. While our method demonstrates excellent performance in reducing missed detections, false positives, and improving localization accuracy across various complex scenes and fog densities, certain limitations remain. The figure illustrates several representative failure scenarios:

\begin{itemize}
    \item The first rows of Fig.~\ref{fig:false_samples} (a) and (b) show that perspective and scene geometry cause background occlusion, distorting target features; even with depth integration, severe occlusion in fog leads to missed detections.
    \item The second rows of Fig.~\ref{fig:false_samples} (a) and (b) show that a drone’s elevated viewpoint shrinks targets, making subtle features indiscernible in fog and resulting in detection failures.
    \item The third rows of Fig.~\ref{fig:false_samples} (a) and (b) show that fog’s scattering induces fading and blurring, distorting critical features and triggering false positives.
\end{itemize}

\begin{figure*}[t]
    \centering
    % \captionsetup{font={footnotesize}}
    \includegraphics[width=0.98\linewidth]{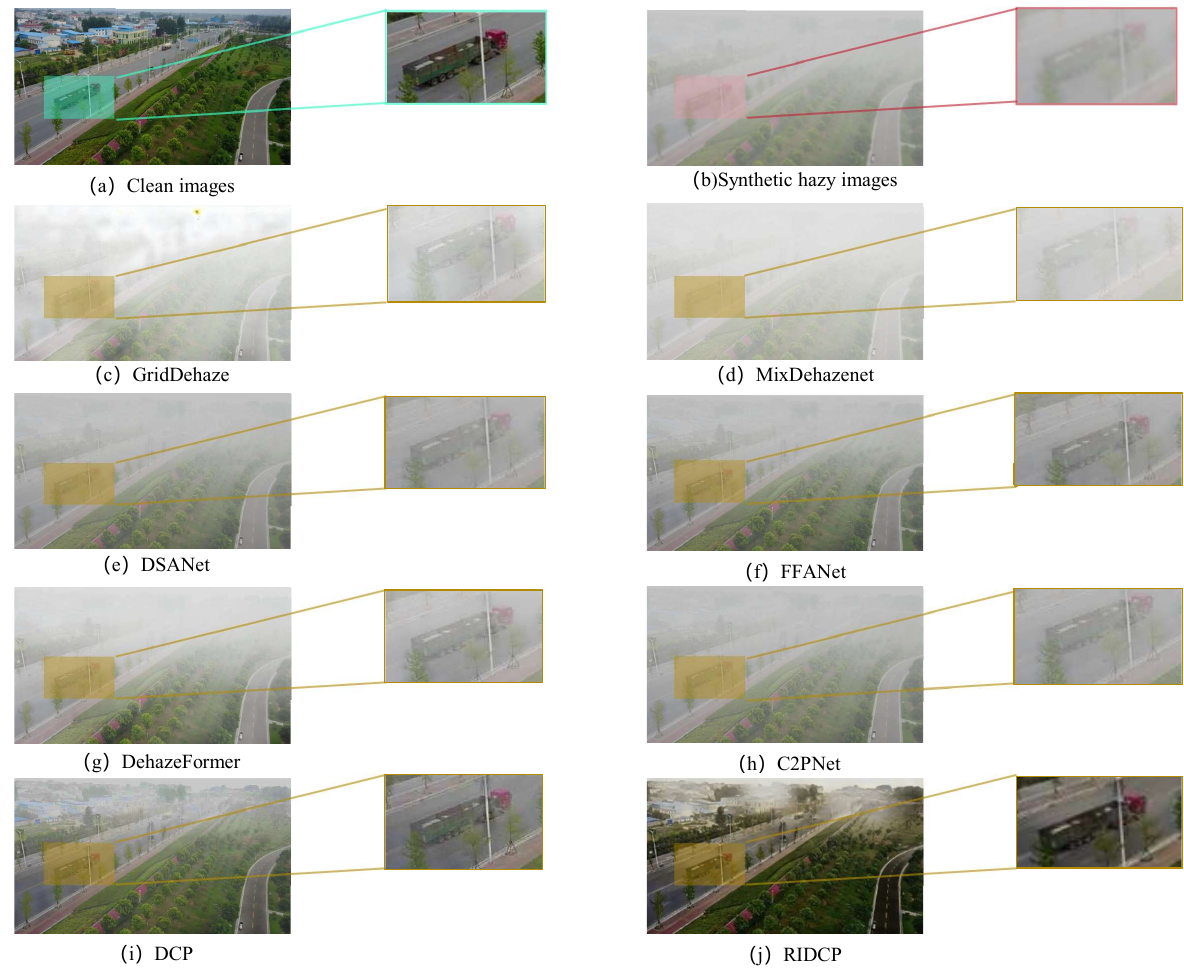}
    \caption{Image dehazing results on HazyDet test-set. From (a) to (j): (a) and (b) show a reference clean image and the corresponding synthetic hazy image, respectively; (c) to (j) are the dehazing outcomes of (c) GridDehaze \citep{ICCV2019GridDehazeNet}, (d) MixDehazeNet \citep{Arxiv2023mixdehazenetmixstructure}, (e) DSANet \citep{NeuralNetworks2024DSANet}, (f) FFANet \citep{AAAI2020FFANet}, (g) DehazeFormer \citep{TIP2023RSHaze}, (h) C2PNet \citep{CVPR2023C2PNet}, (i) DCP \citep{CVPR2009DCP}, (j) RIDCP \citep{CVPR2023RIDCP}, respectively.}
    \label{fig:dehazing_result}
\end{figure*}

\subsection{Dehazing Result Visualizations}\label{subsec:dehaze}

Fig.~\ref{fig:dehazing_result} displays dehazing results from different models. Most dehazing models provided only marginal improvements in objects' clarity and visibility, likely due to inadequate design considerations for UAV perspectives. The relationship between restoration metrics and detection accuracy is complex; higher clarity or subjective image quality, as determined by PSNR and SSIM, does not necessarily improve detection capabilities. Detection models appear to benefit more from preprocessing methods that enhance visual quality while preserving or improving critical features for object detection.

\section{Limitations and Future Work} \label{sec:limit}

In our research, we have identified the considerable potential of simulation data to enhance performance on few-shot visual tasks, particularly for scarce datasets such as drone images captured under adverse conditions, where UAV operation is more challenging than ground vehicle deployment. Following a comprehensive review of existing simulation methods, we proposed a more realistic simulation pipeline. However, inherent distribution differences between simulated and real data persist, potentially affecting model performance in real-world scenarios. In future work, we aim to explore more efficient simulation approaches and extend their application to broader scenarios—such as motion blur and low-light conditions, as well as additional downstream tasks. Furthermore, this paper presents a practical and effective method that leverages depth information as an auxiliary tool for detection tasks. Nonetheless, the performance of the current approach is limited by the accuracy of pseudo-depth labels. Future efforts will focus on incorporating more precise depth data and designing dedicated network architectures to enhance drone object detection under adverse weather conditions.

\section*{Broader Impacts}
\label{sec:broim}

Our progressive fine-tuning approach for adapting visual models from normal to hazy environments offers significant societal benefits by enhancing safety-critical systems' reliability in adverse weather conditions. By bridging the gap between simulated and real-world hazy data, this method substantially reduces the need for expensive and time-consuming collection of real-world hazy imagery, democratizing access to weather-robust AI technologies for research teams with limited resources. The framework's adaptability extends beyond haze to other adverse conditions (fog, rain, snow) and potentially to adjacent fields facing domain gaps such as medical imaging and industrial inspection. While acknowledging potential limitations regarding dataset biases and simulation fidelity, the methodology's contribution to more reliable visual perception systems in challenging environments represents a valuable advancement for autonomous vehicles, surveillance systems, and intelligent infrastructure deployments that must maintain consistent performance regardless of environmental conditions.

\section*{Safeguards}
Our paper employs the ImageNet-21K dataset for pretraining in an open-source multimodal model. Potential security concerns may arise from biases in the pretraining of open-source data and multimodal models. Please be mindful of biases in the original data and model, as well as the security of the model.

%% file: tables/decodet_compare.tex
\begin{table}[th!]  
    \centering  
    % \captionsetup{font={scriptsize}}  
    \caption{\textbf{Comparative experiments of different settings in DeCoDet.}  Bold values indicate the highest performance. mAP--Test and mAP--Real denote average precision on synthetic Test-set and real-world Test-set.}  
    \label{tab:depth_and_noise}  
    \begin{tabular}{@{}ccc@{}}

    \begin{subtable}[t]{0.48\linewidth} 

        \resizebox{\linewidth}{!}  
        {  

        \begin{tabular}{@{}p{3.8cm}*{2}{c}@{}}  
            \toprule  
            Loss function & mAP--Test & mAP--Real  \\   
            \midrule
            L1 loss          & 43.2     & 30.2   \\
            SmoothL1         & 48.9     & 34.3   \\
            MSE Loss         & 49.9     & 36.0   \\
            SIR Loss         & \textbf{52.0}   & \textbf{38.7} \\
            \bottomrule  
        \end{tabular}  
        }  
        \caption{Effectiveness of different depth losses.}  
    \end{subtable} &
    
    \begin{subtable}[t]{0.48\linewidth} 
    
        \resizebox{\linewidth}{!}  
        {  
        \begin{tabular}{@{}l*{2}{c}@{}}  
            \toprule  
            Model & mAP--Test & mAP--Real  \\   
            \midrule
            VA-DepthNet \citep{ICLR2023vadepthnet}        & 43.2     & 30.2   \\
            ZoeDepth \citep{Arxiv2023zoedepth}          & 47.3     & 32.3   \\
            IEBins \citep{NEURIPS2023IEBins}            & 48.5    & 32.1   \\
            UniDepth \citep{CVPR2024Unidepth}           & 49.3     & 34.3   \\
            Metric3D \citep{Arxiv2024metric3d}               & \textbf{52.0}     & \textbf{38.7}   \\
            \bottomrule  
        \end{tabular}  
        }  
        \caption{Effectiveness of different depth maps.}  
    \end{subtable} &
    
    \end{tabular}
    
\end{table}